%% file: main.tex
\documentclass{report}
\PassOptionsToPackage{}{natbib}

    \usepackage[final]{neurips_2020}

\usepackage[utf8]{inputenc} 
\usepackage[T1]{fontenc}    
\usepackage{blindtext}
\usepackage[pagebackref=true,breaklinks=true,colorlinks=true,boo kmarks=false,urlcolor=black, allcolors=mydarkblue]{hyperref}
\usepackage{url}            
\usepackage{booktabs}       
\usepackage{amsfonts}       

\usepackage{nicefrac}       
\usepackage{microtype}      
\usepackage{listings}
\usepackage{xcolor}
\usepackage{tikz}
\definecolor{codegreen}{rgb}{0,0.6,0}
\definecolor{codegray}{rgb}{0.5,0.5,0.5}
\definecolor{codepurple}{rgb}{0.58,0,0.82}
\definecolor{backcolour}{rgb}{0.90,0.90,0.70}

\usepackage{caption}
\input{includes_quant}

\usepackage{changepage}

\makeatletter
\newcommand{\printfnsymbol}[1]{%
  \textsuperscript{\@fnsymbol{#1}}%
}
\makeatother

\makeindex

\usepackage{version}

\author{%
  Sangeetha Siddegowda\\ 
  Qualcomm AI Research\thanks{Qualcomm AI Research is an initiative of Qualcomm Technologies, Inc.}\\
  \texttt{ssiddego@qti.qualcomm.com}
  \And
  Marios Fournarakis\\ 
  Qualcomm AI Research\footnotemark[2]\\
  \texttt{mfournar@qti.qualcomm.com}
  \And
  Markus Nagel\\ 
  Qualcomm AI Research\footnotemark[2]\\
  \texttt{markusn@qti.qualcomm.com}
  \And
  Tijmen Blankevoort \\ 
  Qualcomm AI Research\footnotemark[2]\\
  \texttt{tijmen@qti.qualcomm.com}
  \And
  Chirag Patel \\
  Qualcomm AI Research\footnotemark[2]\\
  \texttt{cpatel@qti.qualcomm.com}
  \And
  Abhijit Khobare \\ 
  Qualcomm AI Research\footnotemark[2]\\
  \texttt{akhobare@qti.qualcomm.com} 
 }

\newcommand*{\fullrefnumname}[1]{\hyperref[{#1}]{\autoref*{#1} \nameref*{#1}}}
\begin{document}

\title{Neural Network Quantization with AI Model Efficiency Toolkit (AIMET)\thanks{AI Model Efficiency Toolkit is a product of Qualcomm Innovation Center, Inc.}}
\maketitle

\newpage
\section*{Acknowledgements}
Dipika Khullar, Frank Mayer, Harshita Mangal, Hitarth Mehta, Jilei Hou, Jonathan Shui,
\\
Joseph Soriaga, Kevin Hsieh, Mart van Baalen, Murali Akula, Sandeep Pendyam, Sendil Krishna,
 \\
Shobitha Shivakumar, Sundar Raman, Tianyu Jiang, Yelysei Bondarenko.
 \\ 
\newpage
\begin{abstract}
While neural networks have advanced the frontiers in many machine learning applications, they often come at a high computational cost. Reducing the power and latency of neural network inference is vital to integrating modern networks into edge devices with strict power and compute requirements. Neural network quantization is one of the most effective ways of achieving these savings, but the additional noise it induces can lead to accuracy degradation.

In this white paper, we present an overview of neural network quantization using AI Model Efficiency Toolkit (\href{https://github.com/quic/aimet}{AIMET}). \href{https://github.com/quic/aimet}{AIMET} is a library of state-of-the-art quantization and compression algorithms designed to ease the effort required for model optimization and thus drive the broader AI ecosystem towards low-latency and energy-efficient inference. \href{https://github.com/quic/aimet}{AIMET} provides users with the ability to simulate as well as optimize \href{https://pytorch.org/}{PyTorch} and \href{https://www.tensorflow.org/}{TensorFlow} models. Specifically for quantization, AIMET includes various post-training quantization (PTQ, cf. chapter \ref{ch:PTQ}) and quantization-aware training (QAT, cf. chapter \ref{ch:QAT}) techniques that guarantee near floating-point accuracy for 8-bit fixed-point inference. We provide a practical guide to quantization via \href{https://github.com/quic/aimet}{AIMET} by covering PTQ and QAT workflows, code examples and practical tips that enable users to efficiently and effectively quantize models using \href{https://github.com/quic/aimet}{AIMET} and reap the benefits of low-bit integer inference.

\end{abstract}

\newpage
\tableofcontents

\title{Introduction}
\input{sections/01_introduction}

\renewcommand{\lstlistingname}{Code Block}
\input{sections/02_quantization_fundamentals}
\input{sections/03_quantization_sim}

\input{sections/04_ptq}

\input{sections/05_qat}

\input{sections/06_conclusions}

\newpage
\input{sections/07_aimet_spec}
\bibliographystyle{icml2020}
\bibliography{refactored_references}

\cleardoublepage
\end{document}

%% file: includes_quant.tex
\usepackage{mathtools}  
\usepackage{caption}
\usepackage{subcaption}  
\usepackage{listings}   
\usepackage{booktabs}   
\usepackage{upgreek}
\usepackage[para]{footmisc}

\newcommand\mat[1]{\mathbf{#1}}

\newcommand\matq[1]{\mat{\widehat{#1}}}

\renewcommand\vec[1]{\mathbf{#1}}

\newcommand{\clamp}{\operatornamewithlimits{clamp}}

\newcommand\ct[1]{#1}

\newcommand\eop[1]{\mathop{\mathbb{E}}\left[#1\right]}

\DeclarePairedDelimiter\round{\lfloor}{\rceil}

\renewcommand{\l}{\left}
\renewcommand{\r}{\right}



%% file: sections/01_introduction.tex
\chapter{Introduction}

With the rise in popularity of deep learning as a general-purpose tool to inject intelligence into electronic devices, the necessity for small, low-latency and energy-efficient neural networks solutions has increased. Today neural networks can be found in many electronic devices and services, from smartphones, smart glasses and home appliances to drones, robots and self-driving cars. These devices are typically subject to strict time restrictions on the execution of neural networks or stringent power requirements for long-duration performance.

One of the most impactful ways to decrease the computational time and energy consumption of neural networks is quantization. In neural network quantization, the weights and activation tensors are stored in lower bit precision than the 16 or 32-bit precision they are usually trained in. When moving from 32 to 8 bit integer, the memory overhead of storing tensors decreases by a factor of 4, while the computational cost for matrix multiplication reduces quadratically by a factor of 16. Neural networks have been shown to be robust to quantization, meaning they can be quantized to lower bit-widths with a relatively small impact on the network's accuracy. Besides, neural network quantization can often be applied along with other common  methods for neural network optimization, such as neural architecture search, compression and pruning. It is an essential step in the model efficiency pipeline for any practical use-case of deep learning. 
However, neural network quantization is not free. Low bit-width quantization introduces noise to the network that can lead to a drop in accuracy. While some networks are robust to this noise, other networks require extra work to exploit the benefits of quantization. 

In literature, quantization issues are handled by using post-training quantization (PTQ) and quantization-aware Training (QAT) techniques. Some of these options are natively supported by frameworks, such as PyTorch and TensorFlow. To make neural network quantization easier for the broader AI ecosystem and enable energy-efficient, fixed-point inference, Qualcomm Innovation Center has open-sourced the AI Model Efficiency Toolkit or \href{https://github.com/quic/aimet}{AIMET}. AIMET is a library of state-of-the-art neural network quantization and compression techniques based on the work of Qualcomm AI Research in this space. This paper provides a practical guide to quantization using AIMET to equip users with sufficient knowledge to quantize their neural networks without requiring in-depth expertise in the domain. For an in-depth discussion of quantization, we encourage readers to refer to our white paper on neural network quantization \citep{whitepaper}. And, documentation corresponding to compression APIs supported by AIMET is available on GitHub at \href{https://github.com/quic/aimet}{https://github.com/quic/aimet}.

\section{AIMET : AI Model Efficiency Toolkit}
\label{sec:aimet}
\begin{figure}[!h]
    \centering
    \includegraphics[width=0.8\textwidth]{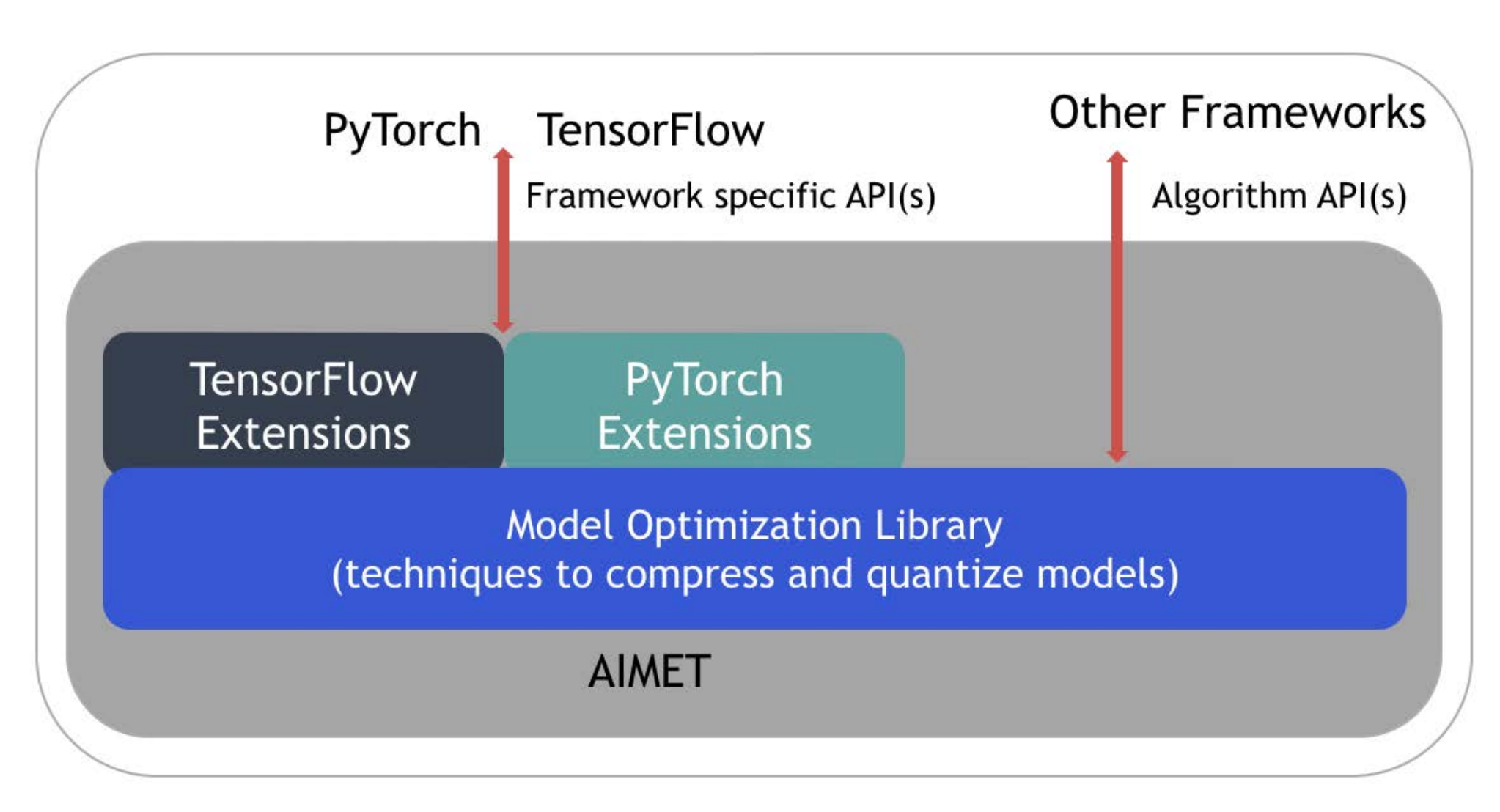}
    \caption{High-level view of AIMET architecture.}
    \label{fig:aimet_high_level}
\end{figure}

A high-level architecture view of AIMET is shown in figure \ref{fig:aimet_high_level}. AIMET implements advanced model quantization and compression algorithms for trained neural network models. It offers easy-to-use APIs to features that have been proven to improve the run-time performance of deep learning neural network models with lower compute and memory requirements and minimal impact on task accuracy. AIMET APIs are designed to work with PyTorch and TensorFlow frameworks. However, the backend stack of the core algorithms is implemented in C++ and compiled into the Model Optimization library. This design allows AIMET to be used as a pluggable toolkit that can be extended to work with other custom frameworks.

\section{AIMET Benefits}
\subsection{Performance}
As mentioned earlier, quantized inference is significantly faster than floating-point inference, but may come at the cost of reduced accuracy. AIMET addresses this by providing a range state-of-the-art quantization techniques from PTQ and QAT literature (more in chapters \ref{ch:PTQ} and \ref{ch:QAT}). The performance of AIMET's quantization techniques has been validated on a wide range of models and tasks (refer to results for PTQ in section \ref{sec:aimet_ptq_results} and QAT in section \ref{sec:QAT_Results}). These optimized models are made available on GitHub as AIMET Model Zoo [\href{https://github.com/quic/aimet-model-zoo}{https://github.com/quic/aimet-model-zoo}]. \footnote{AIMET Model Zoo is a project maintained by Qualcomm Innovation Center, Inc.}

\subsection{Scalability}
Manual optimization of a neural network for improved efficiency is costly, time-consuming and not scalable with ever-increasing AI workloads. AIMET solves this by providing a library that plugs directly into TensorFlow and PyTorch training frameworks for ease of use, allowing developers to call APIs directly from their existing pipelines.

\section{Outline}
In the rest of the paper, we provide a high-level overview of neural network quantization and how AIMET can be used for quantizing neural networks. We start with an introduction to quantization and discuss hardware and practical considerations. We then consider two different regimes of quantizing neural networks: post-training quantization and quantization-aware training. For each regime, we provide AIMET workflows, including code examples, requirements and highlight practical tips to aid users to how to best leverage AIMET for meeting neural network quantization needs.  
Please, note that for brevity, in this guide we only demonstrate code examples\footnote{The code examples covered in this version of the document correspond to AIMET version \href{https://github.com/quic/aimet/releases/tag/1.17.0.py37}{1.17.0.py37}.} from PyTorch APIs. Equivalent TensorFlow APIs can be found in the AIMET documentation available on GitHub at \href{https://quic.github.io/aimet-pages/index.html}{https://github.com/quic/aimet}.

%% file: sections/02_quantization_fundamentals.tex
\chapter{Quantization Fundamentals}

In this chapter, we introduce the basic principles of neural network quantization and of fixed-point accelerators on which quantized networks run on. We start with a hardware motivation and then introduce standard quantization schemes and their properties. Later, we discuss the practical considerations related to layers commonly found in modern neural networks and their implications for fixed-point accelerators.

\section{Hardware background}
\label{sec:hardware_background}
\begin{figure}[h]
\centering
\includegraphics[width=0.7\textwidth]{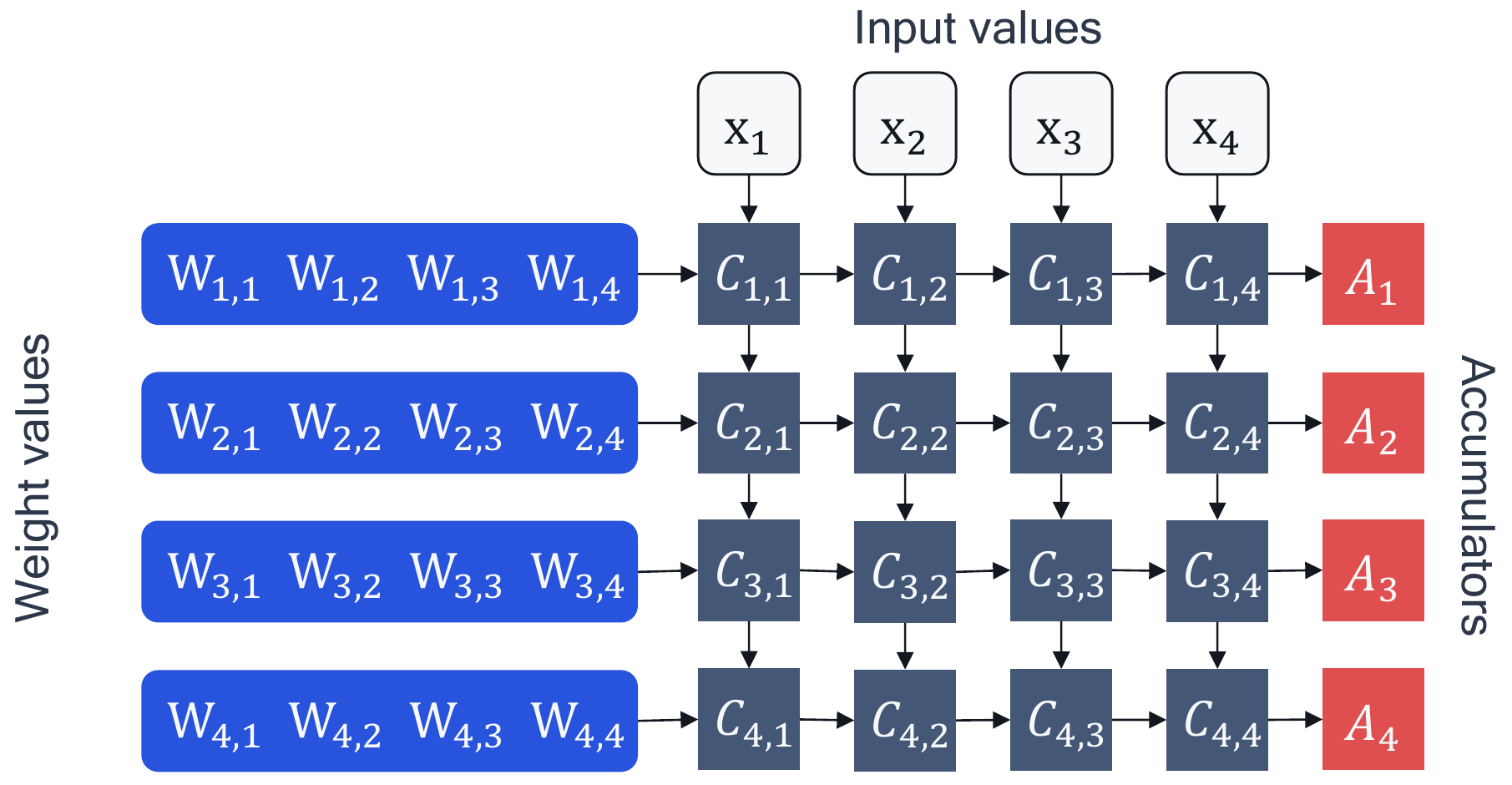}
\caption{A schematic overview of matrix-multiply logic in neural network accelerator hardware.}
\label{fig:nn_accelerator}
\end{figure}

Before diving into the technical details, we first explore the hardware background of quantization and how it enables efficient inference on device. Figure ~\ref{fig:nn_accelerator} provides a schematic overview of how a matrix-vector multiplication, $\mat{y} = \mat{W} \vec{x} + \vec{b}$, is calculated in a neural network (NN) accelerator. This is the building block of larger matrix-matrix multiplications and convolutions found in neural networks. Such hardware blocks aim at improving the efficiency of NN inference by performing as many calculations as possible in parallel.
The two fundamental components of this NN accelerator are the \textit{processing elements} $\ct{C}_{n,m}$ and the \textit{accumulators} $\ct{A}_n$. Our example in figure \ref{fig:nn_accelerator} has 16 processing elements arranged in a square grid and $4$ accumulators. The calculation starts by loading the accumulators with the bias value $\vec{b}_n$.  We then load the weight values $\mat{W}_{n,m}$ and the input values $\vec{x}_m$ into the array and compute their product in the respective processing elements $\ct{C}_{n,m}=\mat{W}_{n,m} \,  \vec{x}_m$ in a single cycle. Their results are then added in the accumulators:
\begin{equation}
    \label{eq:accumulation}
    \ct{A}_n =\vec{b}_n + \sum_{m}{\ct{C}_{n,m}}
\end{equation}
The above operation is also referred to as \textit{Multiply-Accumulate} (MAC). This step is repeated many times for larger matrix-vector multiplications. Once all cycles are completed, the values in the accumulators are then moved back to memory to be used in the next neural network layer. 
Neural networks are commonly trained using FP32 weights and activations. If we were to perform inference in FP32, the processing elements and the accumulator would have to support floating-point logic, and we would need to transfer the 32-bit data from memory to the processing units.
MAC operations and data transfer consume the bulk of the energy spent during neural network inference. Hence, significant benefits can be achieved by using a lower bit fixed-point or \textit{quantized} representation for these quantities. Low-bit fixed-point representations, such as INT8, not only reduce the amount data transfer but also the size and energy consumption of the MAC operation \citep{horowitz}. This is because the cost of digital arithmetic typically scales linearly to quadratically with the number of bits used and because fixed-point addition is more efficient than its floating-point counterpart \citep{horowitz}. Compared to FP32, 8-bit inference provides up to 16x higher performance per watt and thus provides significant energy efficiency for edge inference.

To move from floating-point to the efficient fixed-point operations, we need a scheme for converting floating-point vectors to integers. A floating-point vector $\vec{x}$ can be expressed approximately as a  scalar multiplied by a vector of integer values:
\begin{equation}
\label{eq:simple_quant_example}
\widehat{\vec{x}}  = \ct{s}_{\vec{x}} \cdot \vec{x}_{\text{int}} \approx  \vec{x}
\end{equation}
where $\ct{s}_{\vec{x}}$ is a floating-point \textit{scale factor} and  $\vec{x}_{\text{int}}$ is an integer vector, e.g., INT8. We denote this \textit{quantized} version of the vector as $\widehat{\vec{x}}$. By quantizing the weights and activations we can write the quantized version of the accumulation equation:
\begin{align}
  \hat{\ct{A}}_n& = \widehat{\vec{b}}_n +  \sum_m{\widehat{\mat{W}}_{n,m} \, \widehat{\vec{x}}_m} \nonumber \\
  & =  \widehat{\vec{b}}_n +  \sum_m{ \l(\ct{s}_{\vec{w}} \mat{W}^{\text{int}}_{n,m} \r)\l(\ct{s}_{\vec{x}}\vec{x}^{\text{int}}_m \r)} \nonumber \\
  & =  \widehat{\vec{b}}_n +   \ct{s}_{\vec{w}} \ct{s}_{\vec{x}} \sum_{m}{\mat{W}^{\text{int}}_{n,m} \,\vec{x}^{\text{int}}_m}  \label{eq:quantized_accumulation} 
\end{align}
Note that we used a separate scale factor for weights, $ \ct{s}_{\vec{w}}$, and activations, $\ct{s}_{\vec{x}}$. This provides flexibility and reduces the quantization error (more in section \ref{sec:quant_scheme}). Since each scale factor is applied to the whole tensor, this scheme allows us to factor the scale factors out of the summation in equation \eqref{eq:quantized_accumulation} and perform MAC operations in fixed-point format. We intentionally ignore bias quantization for now, because the bias is normally stored in higher bit-width (32-bits) and its scale factor depends on that of the weights and activations \citep{jacob2018cvpr}.

\begin{figure}[h]
\centering
\includegraphics[width=0.9\textwidth]{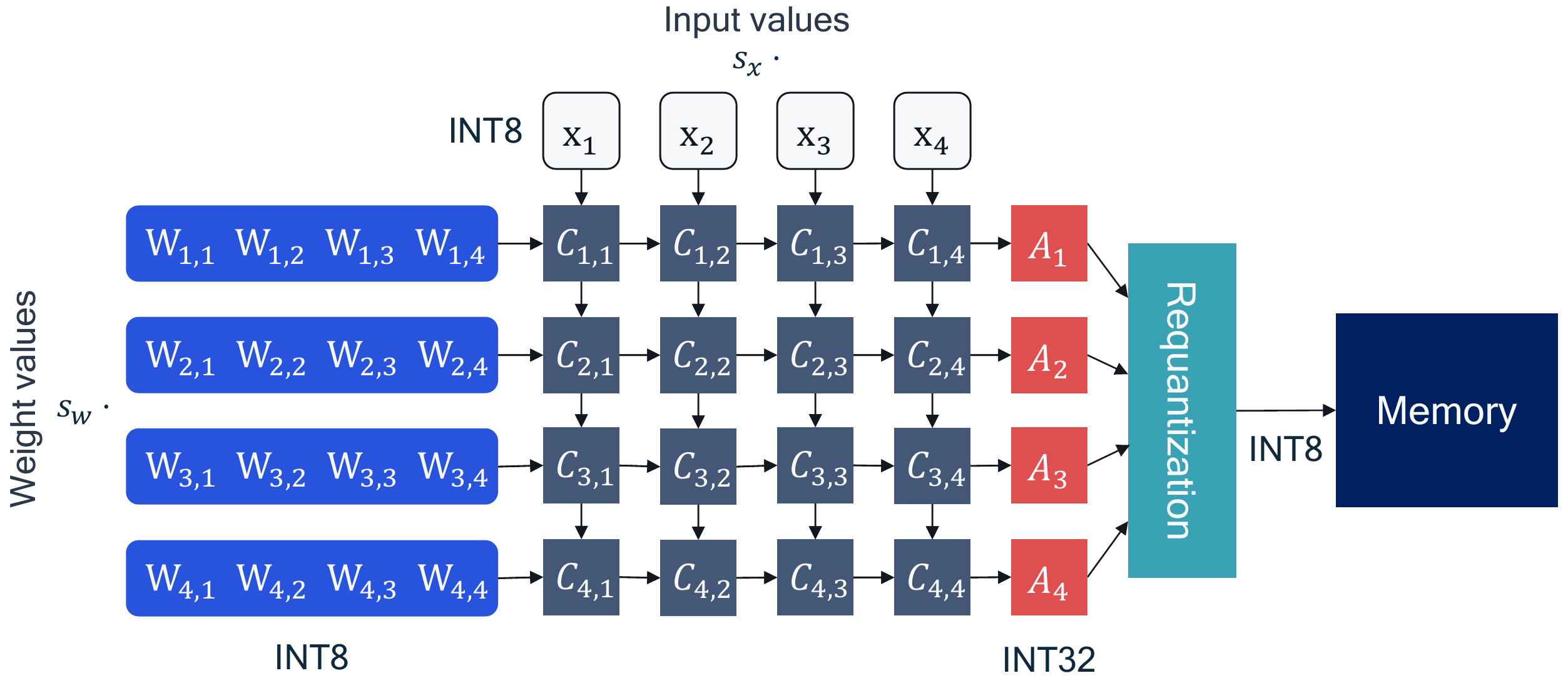}
\caption{A schematic of matrix-multiply logic in an neural network accelerator for quantized inference.}
\label{fig:quanthardware}
\end{figure}

Figure \ref{fig:quanthardware} shows how the neural network accelerator changes when we introduce quantization. In our example, we use INT8 arithmetic, but this could be any quantization format for the sake of this discussion. It is important to maintain a higher bit-width for the accumulators, typical 32-bits wide. Otherwise, we risk incurring loss due to overflow as more products are accumulated during the computation.

The activations stored in the 32-bit accumulators need to be written to memory before they can be used by the next layer. To reduce data transfer and the complexity of the next layer's operations, these activations are quantized back to INT8. This requires a \textit{requantization} step which is  shown in figure \ref{fig:quanthardware}.
 
\section{Concepts}
\paragraph{\textbf{Uniform affine quantization}}
\label{sec:quant_scheme}
\textit{Uniform quantization} is the most commonly used quantization scheme because it permits efficient implementation of fixed-point arithmetic.
\textit{Uniform affine quantization}, also known as \textit{asymmetric quantization}, is defined by three quantization  parameters: the \textit{scale factor} $ 
\ct{s}$, the \textit{zero-point} $\ct{z}$ and the \textit{bit-width} $\ct{b}$. These quantization parameters are also sometimes referred to as \textit{quantization encodings}. A full set of quantization parameters defines a \textit{quantizer}. The scale factor and the zero-point are used to to map a floating point value to the integer grid, whose size depends on the bit-width. The scale factor is commonly represented as a floating-point number and specifies the quantization \textit{step-size}. The zero-point is an integer that ensures that real zero is quantized without error. This is important to ensure that common operations like zero padding or ReLU do not induce quantization error.

Once the three quantization parameters are defined we can proceed with the quantization operation. Starting from a real-valued vector $\vec{x}$ we first map it to the \textit{unsigned} integer grid $\{0,\dots,  2^\ct{b}-1\}$:
\begin{equation}
\label{eq:quant_operation}
    \vec{x}_{\text{int}} =\clamp{\left(\round*{\frac{\vec{x}}{\ct{s}}} + \ct{z} ;0, 2^\ct{b}-1 \right)},
\end{equation}
where $\round*{\cdot}$ is the round-to-nearest operator and clamping is defined as:
\begin{equation}
\label{eq:clamping}
\clamp\left(\ct{x}; \ct{a}, \ct{c}\right) =
    \begin{cases}
            \ct{a}, & \quad \ct{x} <  \ct{a}, \\
            \ct{x}, &  \quad  \ct{a}\leq x  \leq \ct{c},\\
             \ct{c}, & \quad x> \ct{c}.
    \end{cases}
\end{equation}
To approximate the real-valued input $\vec{x}$ we perfrom a  \textit{de-quantization} step:
\begin{equation}
    \label{eq:de_quantization}
        \vec{x}\approx \widehat{\vec{x}}  = \ct{s}\l(\vec{x}_{\text{int}}-\ct{z}\r)
\end{equation}
Combining the two steps above we can provide a general definition for the \textit{quantization function}, $q(\cdot)$, as:
\begin{equation}
    \label{eq:quant_function}
    \widehat{\vec{x}}= q(\vec{x};\ct{s},\ct{z}, \ct{b}) = \ct{s}\left[\clamp{\left(\round*{\frac{\vec{x}}{\ct{s}}}+\ct{z};0 ,2^\ct{b}-1  \right)}-\ct{z}\right], 
\end{equation}
Through the de-quantization step, we can also define the quantization grid limits $(\ct{q}_{\text{min}}, \ct{q}_{\text{max}})$ where $\ct{q}_{\text{min}} = -\ct{s}\ct{z}$ and  $\ct{q}_{\text{max}}=\ct{s}(2^\ct{b}-1-\ct{z})$. Any values of $\vec{x}$ that lie outside this range will be clipped to its limits, incurring a \textit{clipping error}. If we want to reduce the clipping error we can expand the quantization range by increasing the scale factor. However, increasing the scale factor leads to increased \textit{rounding error} as the rounding error lies in the range $\l[-\frac{1}{2}\ct{s}, \frac{1}{2}\ct{s}\r]$. In section \ref{sec:range_setting}, we explore in more detail how to choose the quantization parameters to achieve the right trade-off between clipping and rounding errors.

\begin{figure}[h]
\centering
\includegraphics[width=0.85\textwidth]{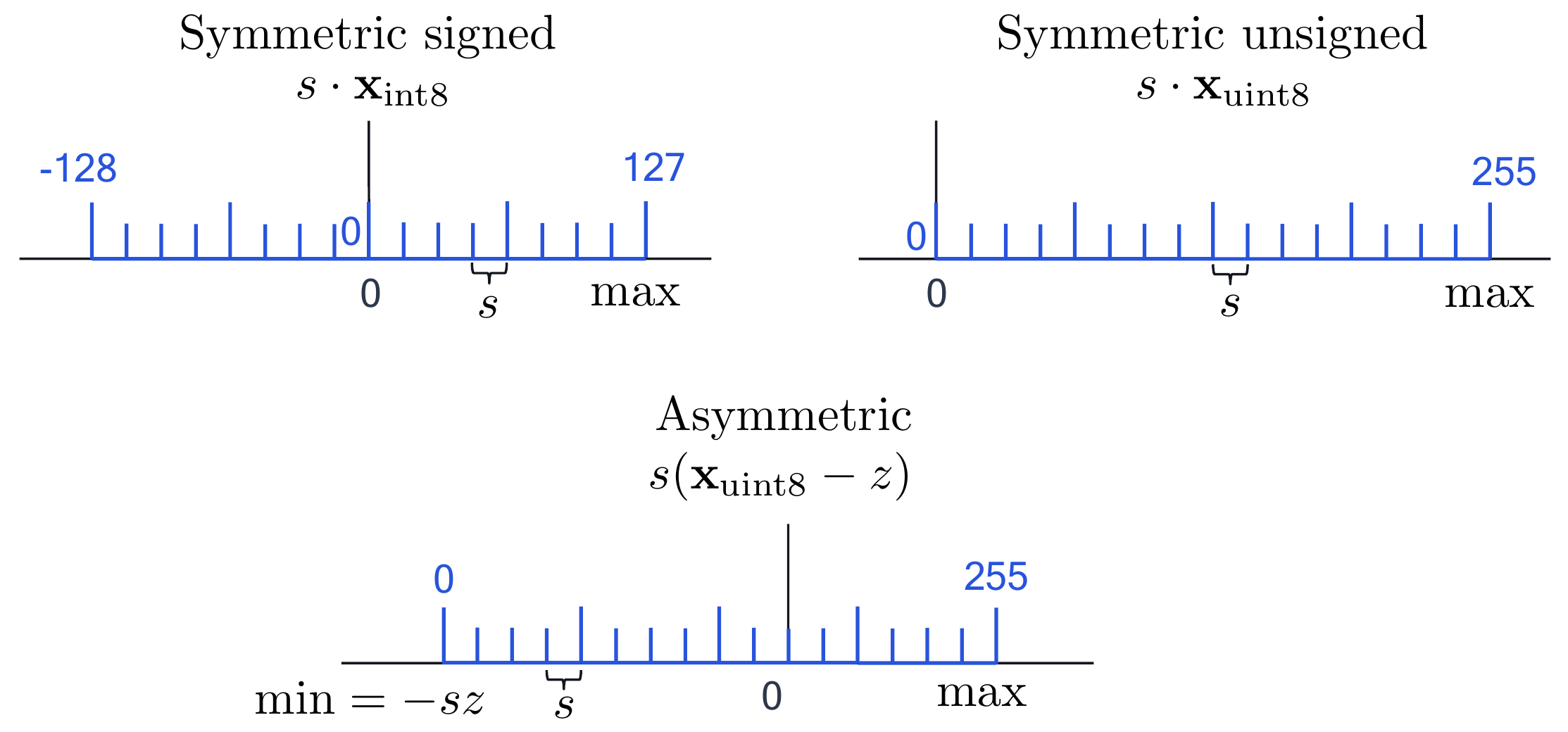}
\caption{A visual explanation of the different uniform quantization grids for a bit-width of 8. $\ct{s}$ is the scaling factor, $\ct{z}$ the zero-point. The floating-point grid is in black, the integer quantized grid in blue.}
\label{fig:quantization_schemes}
\end{figure}

\paragraph{\textbf{Symmetric uniform quantization}}
\label{sec:symmetric_quantizer}
Symmetric quantization is a simplified version of the general asymmetric case. Symmetric quantization restricts the zero-point to 0. This reduces the computational overhead of dealing with zero-point offset during the accumulation operation in equation \eqref{eq:quantized_accumulation}. But the lack of offset restricts the mapping between integer and floating-point domain. As a result, the choice of signed or unsigned integer grid matters:
\begin{subequations}
\label{eq:symmetric_quantization}
\begin{align}
     \widehat{\vec{x}} &= \ct{s} \, \vec{x}_{\text{int}} \\
     \vec{x}_{\text{int}}  &= \clamp\l(\round*{\frac{\vec{x}}{\ct{s}}};0, 2^\ct{b}-1 \r) & \text{for unsigned integers}\\
     \vec{x}_{\text{int}}  &= \clamp\l(\round*{\frac{\vec{x}}{\ct{s}}};-2^{\ct{b}-1} , 2^{\ct{b}-1}-1\r)& \text{for signed integers} 
\end{align}
\end{subequations}
Unsigned symmetric quantization is well suited for one-tailed distributions, such as ReLU activations (see figure \ref{fig:quantization_schemes}). On the other hand, signed symmetric quantization can be chosen for distributions that are roughly symmetric about zero.

\newpage
\paragraph{\textbf{Quantization granularity}}
\label{sec:quant_granularity}
So far, we have defined a single set of quantization parameters, or quantizer, per tensor, one for the weights and one for activations, as seen in equation \eqref{eq:quantized_accumulation}. This is called \textit{per-tensor quantization}. We can also define a separate quantizer for individual segments of a tensor (e.g., output channels of a weight tensor), thus increasing the \textit{quantization granularity}. In neural network quantization, per-tensor quantization is the  the most common choice of granularity due to its simpler hardware implementation: all accumulators in equation \eqref{eq:quantized_accumulation} use the same scale factor, $\ct{s}_{\vec{w}}\ct{s}_{\vec{x}}$. However, we could use finer granularity to further improve performance. For example, for weight tensors, we can specify a different quantizer per output channel. This is known as \textit{per-channel} quantization and its implications are discussed in more detailed in section \ref{sec:per_channel_quantization}. 

Other works go beyond per-channel quantization parameters and apply separate quantizers per group of weights or activations \citep{MSFT,theBitGoesDown,dsConv}.  Increasing the granularity of the groups generally improves accuracy at the cost of some extra overhead. The overhead is associated with accumulators handling  sums of values with varying scale factors. Most existing fixed-point accelerators do not currently support such logic and for this reason, we will not consider them in this work.  However, as research in this area grows, more hardware support for these methods can be expected in the future.

\section {Practical considerations} 
\label{sec:practical considerations}
When quantizing neural networks with multiple layers, we are confronted with a large space of quantization choices including the quantization scheme, granularity, and bit-width. In this section, we explore some of the practical considerations that help reduce the search space.

Note that in this white paper we only consider \textit{homogeneous} bit-width. This means that the bit-width chosen for either weights or activations remains constant across all layers. Homogeneous bit-width is more universally supported by hardware but some recent works also explore the implementation of \textit{heterogeneous} bit-width or \textit{mixed-precision} \citep{bayesianbits,hawq,differentiablequantization}.

\paragraph{\textbf{Symmetric vs. asymmetric quantization}}
\label{sec:symmetric_vs_asymmetric_quant}
For each weight and activation quantization, we have to choose a quantization scheme. On one hand, asymmetric quantization is more expressive because there is an extra offset parameter, but on the other hand there is a possible computational overhead. To see why this is the case, consider what happens when asymmetric weights, $\widehat{\mat{W}} = \ct{s}_{\vec{w}}(\mat{W}_{\text{int}} - \ct{z}
_{\vec{w}})$, are multiplied with asymmetric activations $\widehat{\mat{x}} = \ct{s}_{\vec{x}}(\mat{x}_{\text{int}} - \ct{z}
_{\vec{x}})$:
\begin{align}
    \widehat{\mat{W}} \widehat{\mat{x}} & = \ct{s}_{\vec{w}} (\mat{W}_{\text{int}} - \ct{z}
_{\vec{w}})  \ct{s}_{\vec{x}}(\mat{x}_{\text{int}} - \ct{z}
_{\vec{x}}) \nonumber \\
    & = \ct{s}_{\vec{w}}  \ct{s}_{\vec{x}} \mat{W}_{\text{int}} \mat{x}_{\text{int}}  - \textcolor{red}{\ct{s}_{\vec{w}} \ct{z}
_{\vec{w}}  \ct{s}_{\vec{x}} \mat{x}_{\text{int}}} - \textcolor{blue}{\ct{s}_{\vec{w}}  \ct{s}_{\vec{x}} \ct{z}
_{\vec{x}} \mat{W}_{\text{int}} + \ct{s}_{\vec{w}} \ct{z}
_{\vec{w}} \ct{s}_{\vec{x}} \ct{z}
_{\vec{x}}}. \label{eq:asymmetric_integer_matmult}
\end{align}
The first term is what we would have if both operations were in symmetric format. The third and fourth terms depend only on the scale, offset and weight values, which are known in advance. Thus these two terms can be pre-computed and added to the bias term of a layer at virtually no cost. The second term, however, depends on the input data $\mat{x}$. This means that for each batch of data we need to compute an additional term during inference. This can lead to significant overhead in both latency and power, as it is equivalent to adding an extra channel.

For this reason, it is a common approach to use \textit{asymmetric activation quantization} and \textit{symmetric weight quantization} that avoids the additional data-dependent term. 

\paragraph{\textbf{Per-tensor and per-channel quantization}}
\label{sec:per_channel_quantization}
In section \ref{sec:quant_granularity}, we discussed different levels of quantization granularity. Per-tensor quantization of weights and activations has been standard for a while because it is supported by all fixed-point accelerators. However, per-channel quantization of the weights can improve accuracy, especially when the distribution of weights varies significantly from channel to channel. Looking back at the quantized MAC operation in equation \eqref{eq:quantized_accumulation}, we can see that per-channel weight quantization can be implemented in the accelerator by applying a separate per-channel weight scale factor without requiring rescaling. Per-channel quantization of activations is much harder to implement because we cannot factor the scale factor out of the summation and would, therefore, require rescaling the accumulator for each input channel. Whereas \textit{per-channel quantization} of weights is increasingly becoming common practice, not all commercial hardware supports it. Therefore, it is important to check if it is possible in your intended target device.


%% file: sections/03_quantization_sim.tex
\chapter{Quantization Simulation}

To test how well a neural network would run on a quantized device, we often simulate the quantized behavior on the same general-purpose machine we use for training neural networks. This is called \textit{quantization simulation}. 

\section{Introduction}
\label{sec:quantsim_intro}

\begin{figure} [!h]
     \centering
     \begin{subfigure}[b]{0.46\textwidth}
         \centering
         \includegraphics[width=0.75\textwidth]{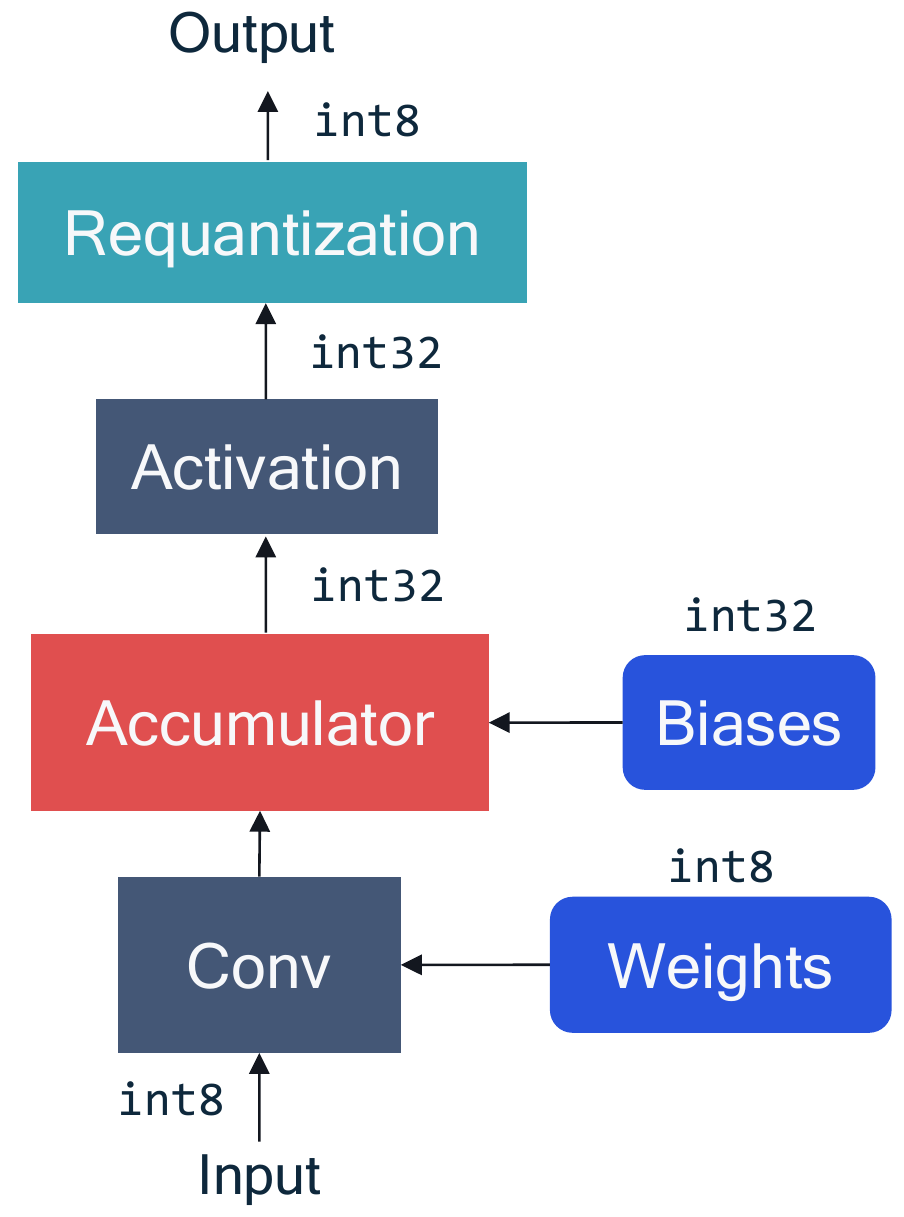}
         \caption{Diagram for quantized on-device inference with fixed-point operations.}
         \label{fig:on-device_quant_inference}
     \end{subfigure}
     \hfill
     \begin{subfigure}[b]{0.44\textwidth}
         \centering
         \includegraphics[width=\textwidth]{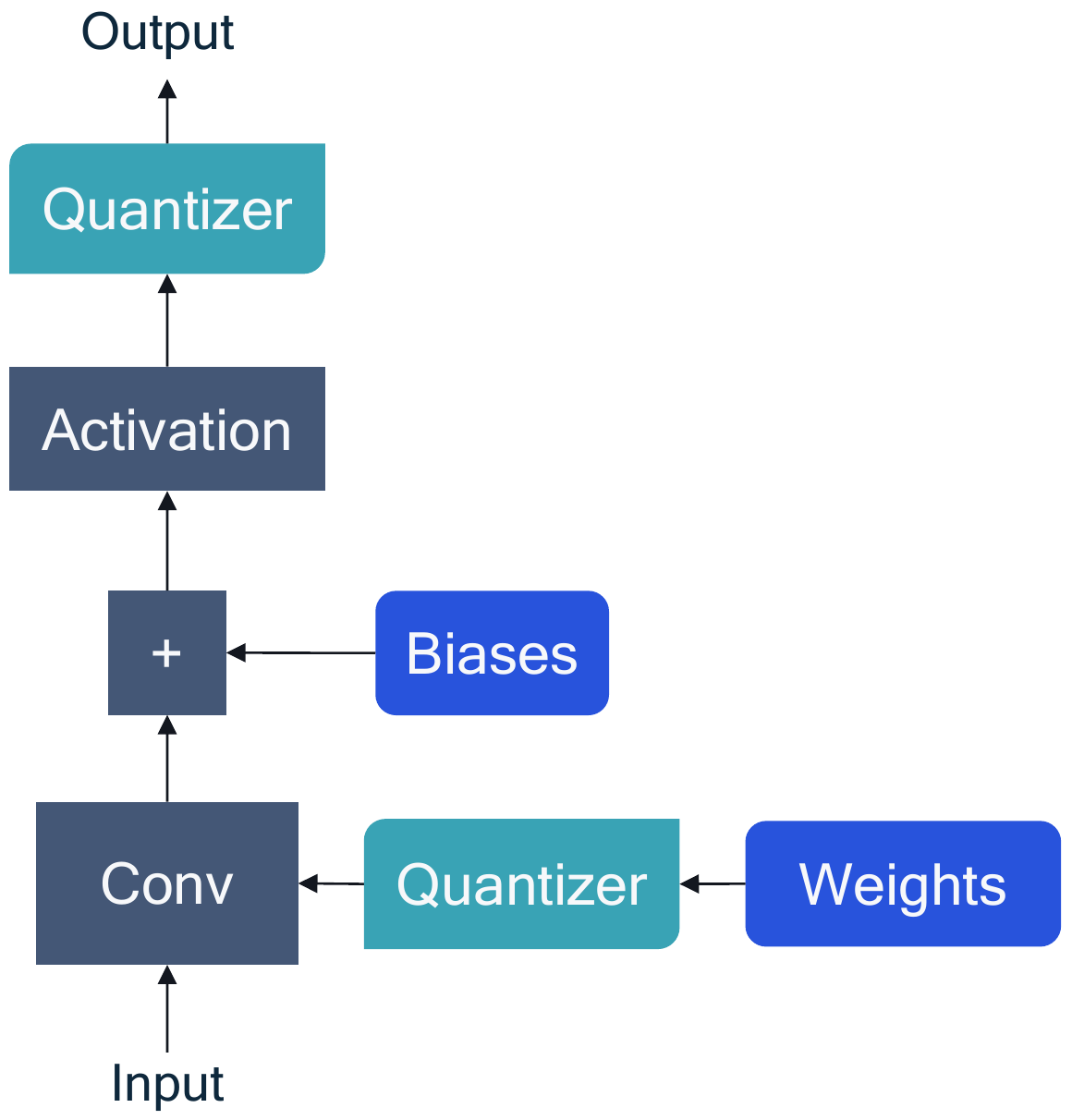}
         \caption{Simulated quantization using floating-point operations.}
         \label{fig:simulated quantization}
     \end{subfigure}
        \caption{Schematic overview of quantized forward pass for a convolutional layer.}
        \label{fig:simulation}
\end{figure}

Quantization simulation allows the user to efficiently test various quantization options. In this section, we first explain the fundamentals of this simulation process and then discuss techniques that help bridge the gap between simulated and the actual on-device performance.

AIMET enables a user to automatically create a quantization simulation model, given a PyTorch or TensorFlow model. AIMET will add the \textit{quantizer} nodes shown in figure \ref{fig:simulation} to the PyTorch or TensorFlow model graph. The resulting model graph can be used as-is in the user's evaluation or training pipeline. 

AIMET quantization simulation can be use for the following use cases:
\begin{itemize}
    \item Simulate on-device inference without the need for the runtime or target device.
    \item Enable quantization-aware training, as described later in chapter \ref{ch:QAT}.
    \item It is used under the hood for other AIMET quantization features, such as AdaRound (cf. section \ref{sec:adaround}) or bias correction (cf. section \ref{sec:bias_correction}).

\end{itemize}

\paragraph{\textbf{Workflow}}

\begin{figure}[!h]
    \centering
    \includegraphics[width=1\textwidth]{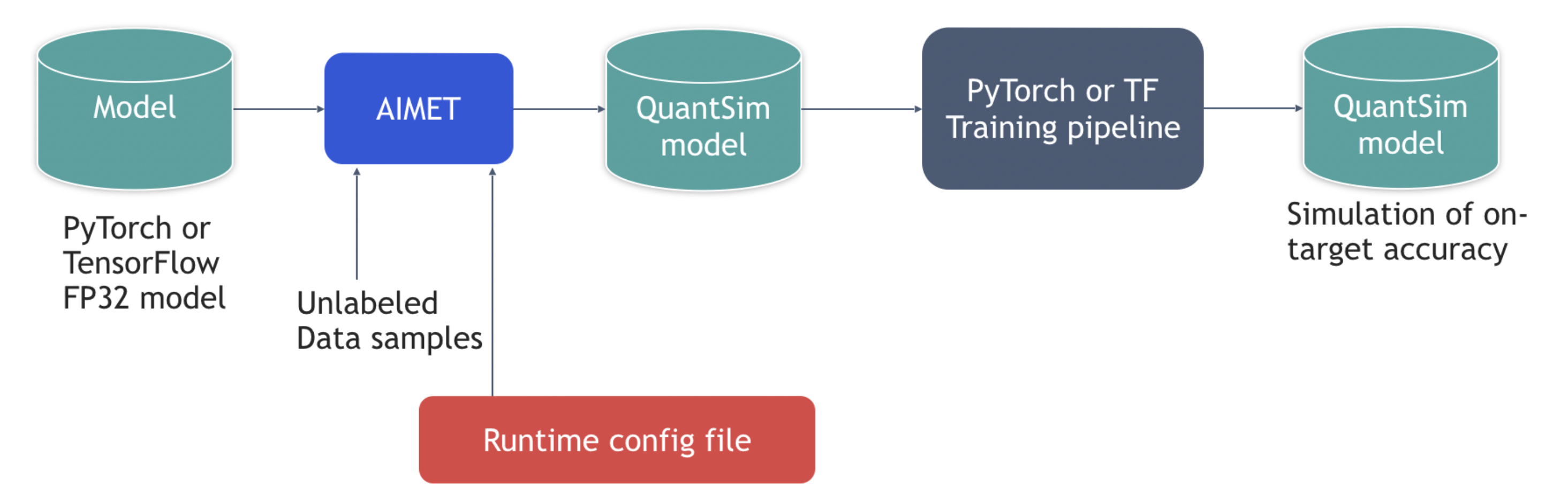}
    \caption{AIMET quantization simulation workflow.}
    \label{fig:aimet_quantization_sim_workflow}
\end{figure}

Figure \ref{fig:aimet_quantization_sim_workflow} shows the workflow for using AIMET quantization simulation to simulate on-target quantized accuracy.

\begin{itemize}
\item The user starts with a pretrained floating-point FP32 model.
\item AIMET creates a simulation model by inserting quantization simulation ops into the model graph.
\item AIMET also configures the inserted simulation ops. In section \ref{sec:quantsim_config} we show how the configuration of these ops can be controlled via a configuration file.
\item AIMET finds optimal quantization parameters, such as scale/offsets, for the inserted quantization simulation ops. To this end, AIMET requires the user to provide a callback method that feeds a few representative data samples through the model. These samples can either be from the training or calibration datasets.  Generally, samples in the order of $1{\small,}000$ have been sufficient for AIMET to find optimal quantization parameters.
\item AIMET returns a quantization simulation model that can be used as a drop-in replacement for the original model in their evaluation pipeline. Running this simulation model through the evaluation pipeline yields a \textit{quantized accuracy} metric that closely simulates on-target accuracy.
\end{itemize}

\paragraph{\textbf{Code example}}
\begin{itemize}
    
    \item [\colorbox{gray!40}{API}] A quantization simulation model can be created using a single API call, as shown below. A second call to \texttt{compute\_encodings()} generates the scale/offset parameters for the quantization sim ops. The user can now feed the simulation model (\texttt{sim.model}) to their evaluation pipeline/function, as shown in the the code example below.
    
\begin{lstlisting}[caption={Quantization simulation API example.},language=Python]

    import torch
    from aimet_torch.examples import mnist_torch_model
    # Quantization related import
    from aimet_torch.quantsim import QuantizationSimModel

    model = mnist_torch_model.Net().to(torch.device('cuda'))

    # Create a quantization simulation model
    # Adds simulation ops, configures these simulation ops
    
    sim = QuantizationSimModel(model, 
              dummy_input=torch.rand(1, 1, 28, 28)
              default_output_bw=8, default_param_bw=8)

    # Find optimal quantization parameters (like scale/offset)
    # The callback method should send representative data 
    # samples through the model. The forward_pass_callback_args
    # argument is passed as-is to the callback method
    
    sim.compute_encodings(forward_pass_callback=send_samples, 
                          forward_pass_callback_args=None)

    # We use sim.model as a drop-in replacement for model in 
    # an eval pipeline

    quantized_accuracy = eval_function(model=sim.model)
    \end{lstlisting}
\end{itemize}

\begin{itemize}
    \item [\colorbox{gray!40}{Usage Note}]When using AIMET quantization simulation with PyTorch models:
    \item AIMET quantization simulation requires the model definition to follow certain guidelines, which are detailed in section \ref{sec:pt_model_guidelines}.
    \item AIMET also includes a \textit{Model Validator} tool that allows the user to check their model definition and find constructs that may need replacing. Please check the AIMET documentation on GitHub: \href{https://quic.github.io/aimet-pages/index.html}{https://quic.github.io/aimet-pages/index.html}, to learn more about this tool.
\end{itemize}

\section{Practical considerations}
\label{sec:quantsim_bn_fold}
\paragraph{\textbf{Batch normalization folding (BNF)}}
\label{sec:batch_norm_folding}
 Batch normalization \citep{batchnorm} is a standard component of modern convolutional networks. Batch normalization operation normalizes the output of a linear or convolutional layer before scaling and adding an offset. For on-device inference, these operations are folded into the previous or next convolutional or linear layers in a step called \textit{batch normalization folding} \citep{krishnamoorthi,jacob2018cvpr}. This removes the batch normalization operations entirely from the network, as the calculations are absorbed into an adjacent layer. Besides reducing the computational overhead of the additional scaling and offset, this prevents extra data movement and the quantization of the layer's output.

It is recommended that users fold batch normalization layers prior to using AIMET quantization simulation. This ensures that the simulated quantized accuracy produced using AIMET will closely mimic the on-target inference accuracy. AIMET provides APIs to allow the user to fold batch normalization layers automatically.

AIMET can automatically detect batch normalization layers in the model definition and can fold them into adjacent convolutional layers. This API will fold the batch normalization layers in-place. 

\paragraph{\textbf{Code example}}
\begin{itemize}
\item [\colorbox{gray!40}{API}]
Below is a code example demonstrating using AIMET to fold batch normalization layers.
\begin{lstlisting}[caption={Batch normalization folding API example.},language=Python]
    
    from torchvision import models
    from aimet_torch.batch_norm_fold import fold_all_batch_norms

    model = models.resnet18(pretrained=True).eval()
    
    fold_all_batch_norms(model, input_shape=(1, 3, 224, 224))  
    \end{lstlisting}

\item [\colorbox{gray!40}{Usage Note}]
    Please check the list of currently supported convolution and batch normalization types under section \ref{sec:supported_types}.
\end{itemize}

\section{Exporting quantization encodings}
\label{sec:quantsim_export}
Quantization simulation does not change the model parameters itself. However, as part of creating the simulation model, AIMET computes optimal scale/offset \textit{quantization encodings} for both parameters and activations in the model. These \textit{quantization encodings} can be exported from AIMET. On-target runtimes, such as the Qualcomm Neural Processing SDK, can import these optimized \textit{quantization encodings} and use them instead of computing quantization encodings independently.
    \begin{figure} [!h]
        \centering
        \includegraphics[width=1 \textwidth]{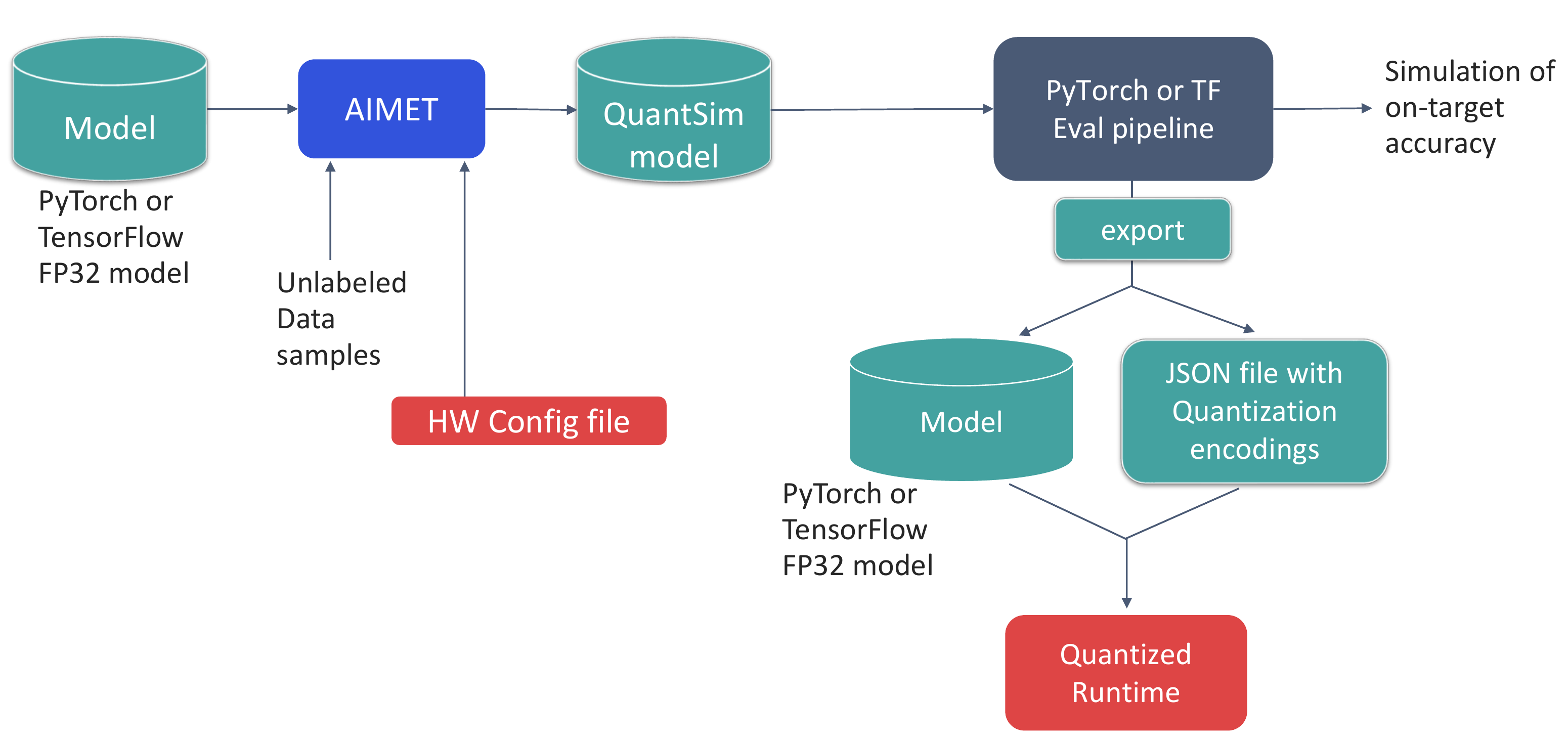}
        \caption{Workflow to export AIMET quantization simulation encodings.}
        \label{fig:aimet_quantization_sim_export}            
    \end{figure}

The above figure shows the workflow for exporting AIMET optimized quantization encodings. After the simulation model is created, AIMET provides an export API that will export both the original FP32 model without the simulation ops and also a JSON-formatted file that contains the optimized quantization encodings. Please refer to AIMET documentation to understand the structure of this encodings file.

\paragraph{\textbf{Code example}}
\begin{itemize}
    
    \item [\colorbox{gray!40}{API}]
    The following code example show how an AIMET quantization simulation model is created and initialized. The \texttt{export()} method generates the following files:
    \begin{enumerate}
        \item A PyTorch model without any simulation ops. This is equivalent to the model initially used to create the quantization simulation model.
        \item Quantization encodings are exported to a separate JSON-formatted file that can then be imported by the on-target runtime (if desired).
        \item An equivalent model in ONNX format is exported. The encodings file above refers to tensor names in the exported ONNX model file.
    \end{enumerate}
    \begin{lstlisting}[caption={Exporting quantization encodings API example.},language=Python]   
    
    import torch
    from aimet_torch.examples import mnist_torch_model
    # Quantization related import
    from aimet_torch.quantsim import QuantizationSimModel

    model = mnist_torch_model.Net().to(torch.device('cuda'))

    # Create a quantization simulation model
    # Adds simulation ops, configures these simulation ops

    sim = QuantizationSimModel(model, 
              dummy_input=torch.rand(1, 1, 28, 28),
              default_output_bw=8, 
              default_param_bw=8)

    # Find optimal quantization parameters (like scale/offset)
    # The callback method should send representative data 
    # samples through the model. The forward_pass_callback_args 
    # argument is passed as-is to the callback method

    sim.compute_encodings(forward_pass_callback=send_samples, 
                          forward_pass_callback_args=None)

    # Optionally, the user can export out the quantization 
    # parameters (like per-layer scale/offset)

    sim.export(path='./', filename_prefix='quantized_mnist', 
               dummy_input=torch.rand(1, 1, 28, 28))    
    \end{lstlisting}
\end{itemize}

\section{Configuration of quantization simulation ops in the model}
\label{sec:quantsim_config}

Different hardware and on-device runtimes may support different quantization choices for neural network inference. For example, some runtimes may support asymmetric quantization for both activations and weights, whereas other ones may support asymmetric quantization just for weights. As a result, we need to make quantization choices during simulation that best reflect our target runtime and hardware.

To this end, AIMET uses a configuration driven approach to set up the simulation ops inserted into the model. The \textit{QuantizationSimModel} API, shown in the previous code example, takes a JSON runtime-configuration file as an optional parameter. This configuration file contains six main sections, in increasing amounts of specificity as shown in figure \ref{fig:aimet_quantization_sim_config}. This configuration file can be tailored to specific hardware configuration.
    \begin{figure} [!h]
        \centering
        \includegraphics[width=0.85 \textwidth]{figures/quantsim_config_file}
        \caption{AIMET quantization simulation configuration.}
        \label{fig:aimet_quantization_sim_config}            
    \end{figure}

Please refer to  the AIMET documentation on \href{https://quic.github.io/aimet-pages/index.html}{ GitHub} for details on usage of this configuration file.

%% file: sections/04_ptq.tex
\chapter{Post-Training Quantization}
\label{ch:PTQ}
\section{Introduction}
Post-training quantization algorithms take a pretrained FP32 network and optimize its weights and quantization parameters for improved quantized performance without any fine-tuning. PTQ methods can be data-free or may require a small calibration set, which is often readily available. Because they do not require access to the original training pipeline and involve almost no hyperparameter tuning, they can be used via a single API call, as a black-box method, to quantize a pretrained neural network in a computationally efficient manner. This frees the neural network designer from being an expert in quantization and thus allows for a much broader application of neural network quantization.

AIMET supports state-of-the-art quantization techniques, such as data-free quantization (DFQ) \citep{dfq} and AdaRound \citep{adaround}. Before discussing their usage, we first present our best-practice PTQ pipeline. 

\section{Standard PTQ pipeline}
\label{sec:standard_ptq}
\begin{figure}[!h]
    \centering
    \includegraphics[width=1.0\textwidth]{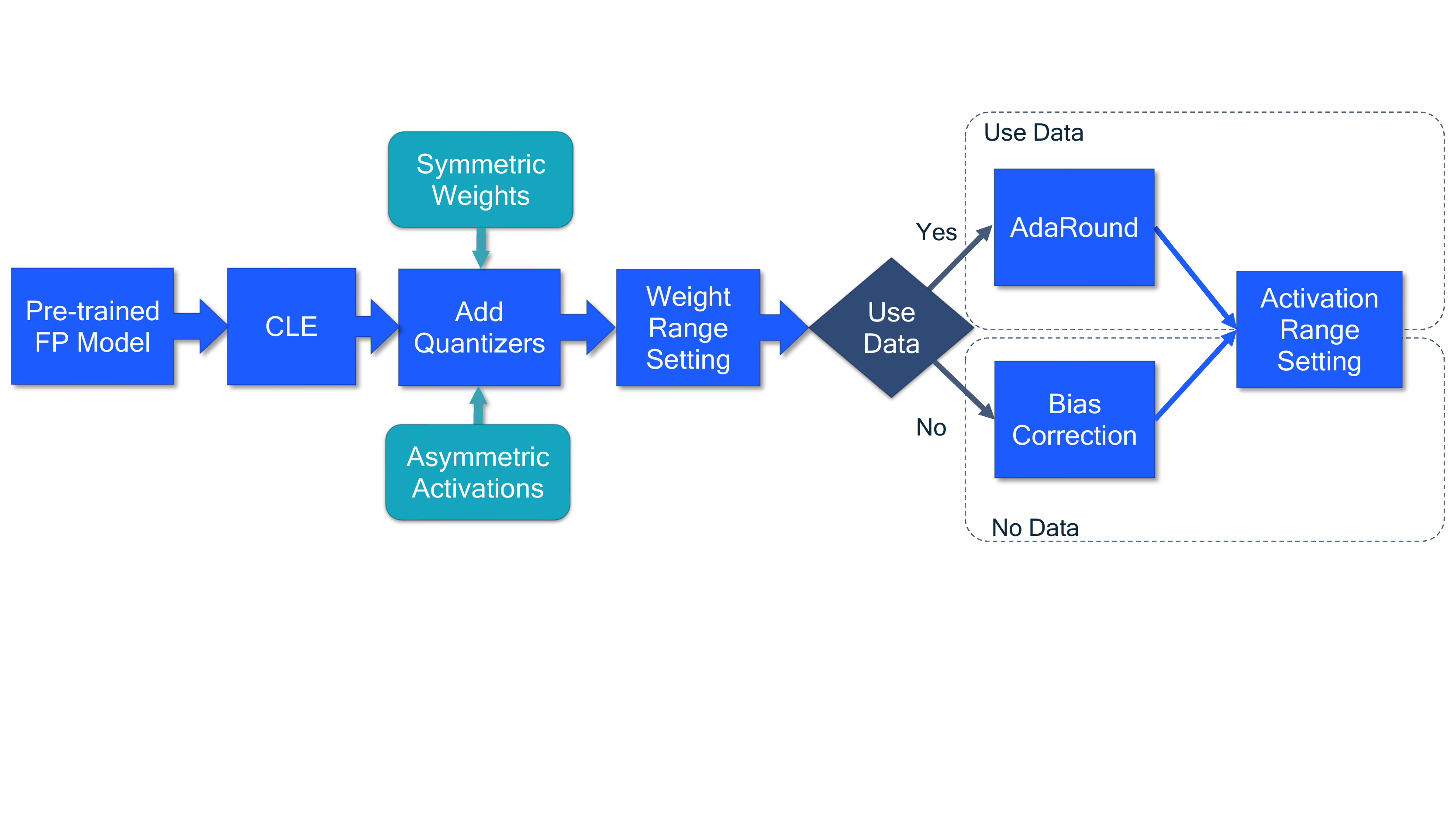}
    \caption{Standard PTQ pipeline. Blue boxes represent required steps and the turquoise boxes recommended choices.}
    \label{fig:standard_ptq}
\end{figure}

The recommended PTQ pipeline is illustrated in figure \ref{fig:standard_ptq}.  This pipeline is based on relevant literature and extensive experimentation and achieves competitive results for many computer vision and natural language processing models and tasks. Depending on the model, some steps may not be required, or other choices can lead to equal or better performance.

\begin{description}
\item[Cross-layer equalization] First we apply cross-layer equalization (CLE), a pre-processing step for the full precision model to make it more quantization friendly. CLE is particularly beneficial for models with depth-wise separable layers when using per-tensor quantization, but it can also improve other layers and quantization choices.

\item[Add quantizers] Next, we choose our quantizers and add quantization operations in our network as detailed in section \ref{sec:quantsim_intro}. The choice of quantizer depends on the specific target HW; for common AI accelerators we recommend using symmetric quantization for the weights and asymmetric quantization for the activations. If supported by the HW/SW stack, then it is favorable to use per-channel quantization for weights.

\item[Weight range setting] Next, we set the quantization parameters of all weight tensors using the quantization range setting method described in section \ref{sec:range_setting}. We recommend using the layer-wise Signal-to-Quantization-Noise (SQNR) based criteria. In the specific case of per-channel quantization, using the min-max method can be favorable in some cases.

\item[AdaRound] If a small calibration dataset\footnote{Usually, between $500$ and $1{\small,}000$ unlabeled images are sufficient as a calibration set.} is available, we then apply AdaRound to optimize the rounding of the weights. This step is crucial to enable low-bit weight quantization (e.g. 4 bits) in PTQ.

\item[Bias correction] If a calibration dataset is not available and the network uses batch normalization, we can use analytical bias correction instead. We refer to this suite of techniques employed under CLE along with bias correction as \textit{data-free quantization}.

\item[Activation range setting] As the final step, we determine the quantization ranges of all data-dependent tensors in the network, such as activation tensors using SQNR based criteria for most of the layers. A small calibration dataset is required to perform this step. 
\end{description}

\section{Cross-layer equalization}
\label{sec:CLE}
A common issue in neural network quantization is that the dynamic range of weight channels can vary significantly. This can lead to poor quantization accuracy, especially, for per-tensor quantization. Cross-layer equalization (CLE) addresses this issue by exploiting the scale equivariance property of certain activation functions (e.g. ReLU, PReLU) to equalize the weight tensor and improve quantization performance. 
AIMET enables applying CLE with a single API call and shows state-of-the-art quantization performance for common computer vision architectures and, particularly, in the family of efficient models, such as MobileNet.

\begin{figure}[ht]
    \centering
    \includegraphics[scale=0.45]{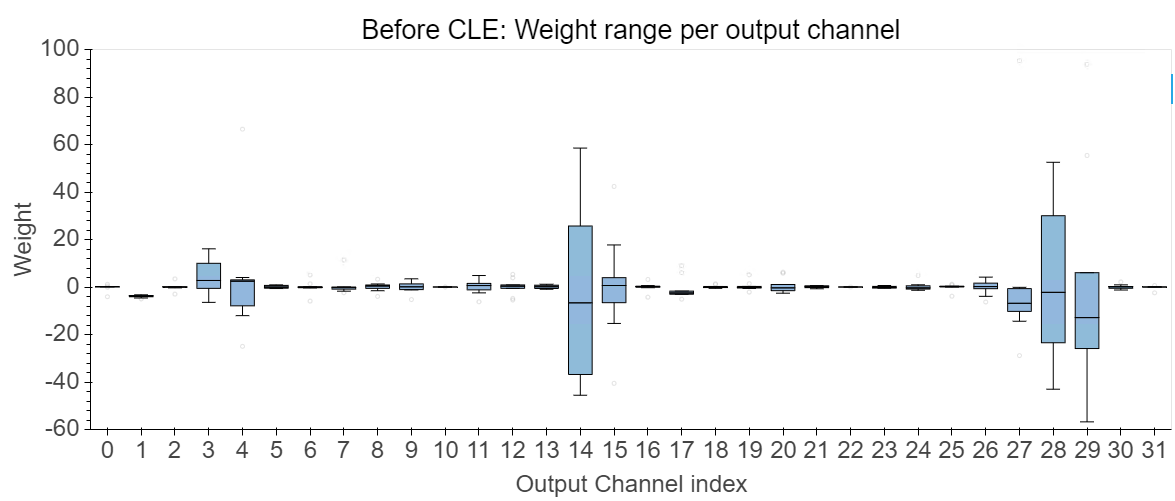}
    \caption{AIMET visualization example: per (output) channel weight ranges of the first depthwise-separable layer in MobileNetV2 after BN folding.
    The boxplots show the min and max value range for each channel before CLE.}
    \label{fig:aimet_viz_before_cle}           
\end{figure}

\begin{figure}[ht]
    \centering
    \includegraphics[scale=0.45]{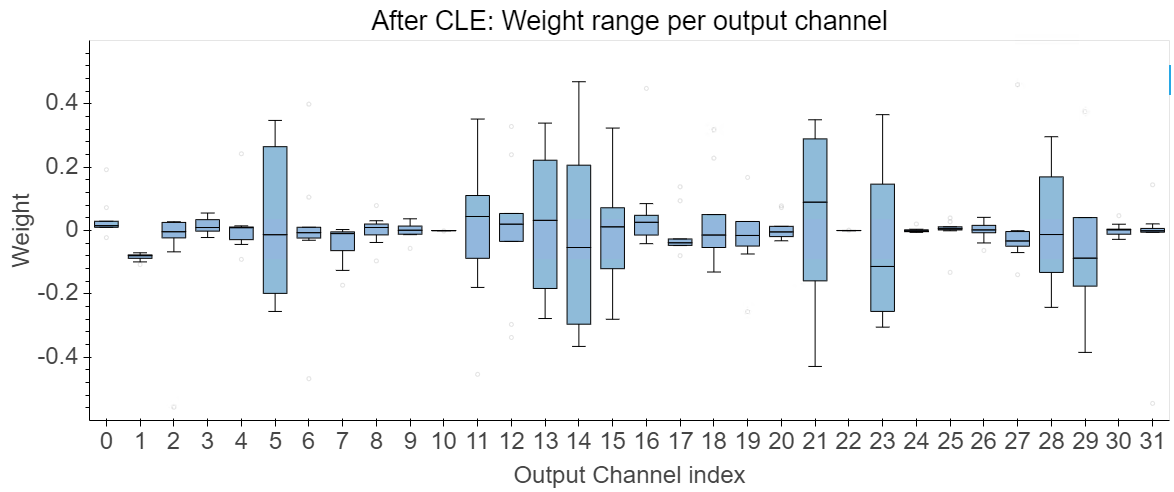}
    \caption{AIMET visualization example: per (output) channel weight ranges of the first depthwise-separable layer in MobileNetV2.
    The boxplots show the min and max value range for each channel after CLE.}
    \label{fig:aimet_viz_after_cle}           
\end{figure}

\paragraph{\textbf{Code example}}
\begin{itemize}
    \item [\colorbox{gray!40}{API}]
    AIMET cross-layer equalization technique can be applied on a given pretrained model using one unified AIMET API shown below.
    \begin{lstlisting}[caption={Unified top-level equalize model API example.},language=Python]
    from torchvision import models
    from aimet_torch.cross_layer_equalization import equalize_model

    model = models.resnet18(pretrained=True).eval()
    input_shape = (1, 3, 224, 224)

    # Performs batch normalization folding, Cross-layer scaling and High-bias absorption and updates model in-place
    equalize_model(model, input_shape) 
    \end{lstlisting}

    \item [\colorbox{gray!40}{Usage Note}]
        \item CLE is particularly beneficial for models with depth-wise separable layers when using per-tensor quantization, but it often also improves other layers and quantization choices.
        \item Please, note that the unified CLE API implements the following steps:
       \begin{enumerate}
       \item batch normalization folding (cf. section \ref{sec:batch_norm_folding}),
       \item replaces ReLU6 activations with RELU.
       However, this replacement could lead to accuracy degradation. (follow guidelines under cf. section \ref{sec:CLE_prereq})
       \item applies CLE for weights equalization (cf. section \ref{sec:CLE}),
       \item and, finally, absorbs high-bias to equalize activations.
       \end{enumerate}
       \item The user should use the original model prior to batch normalization folding as input to this API.
        \item CLE can be applied before quantization simulation (cf. section \ref{sec:quantsim_intro}) but always before range setting.
       \item Before applying CLE, the user can use AIMET to visualize the weight ranges of the model and thus determine if a model is a candidate for applying the CLE technique. For example, if the dynamic range of weights varies significantly across channels of a given convolutional layer, then CLE can be beneficial. Further, visualizations can be used to understand the effect or improvements after applying CLE to a given model. Two sample visualization graphs are shown in figure \ref{fig:aimet_viz_before_cle} and figure \ref{fig:aimet_viz_after_cle}. 

        \item AIMET also supports low-level APIs that enable using the CLE techniques independently. For more details on this and visualization API, please refer to AIMET documentation on GitHub (\href {https://quic.github.io/aimet-pages/index.html}{https://quic.github.io/aimet-pages/index.html}).
\end{itemize}

\newpage
\subsection{Caveats}
\label{sec:CLE_prereq}
We recommend performing a baseline accuracy check with replacement of ReLU6 with ReLU activation on a given model before applying the CLE suite of techniques. If the floating-point accuracy drops on this replacement, do not apply CLE. Instead, an alternate technique, such as AdaRound, detailed in section \ref{sec:adaround} can be explored.
\paragraph{\textbf{Replace ReLU6 with ReLU}}
\label{sec:replace_relu6_with_relu}
AIMET supports a low-level API to replace ReLU6 layers with ReLU activations in a given model. 
The input model passed to the API is modified in-place to replace ReLU6 with ReLU activations.
\paragraph{\textbf{Code example}}
\begin{itemize}
    \item [\colorbox{gray!40}{API}]
    The replacement can be easily performed using generic AIMET API as shown below.
    \begin{lstlisting}[caption={ReLU6 replacement API example.},language=Python]

    from torchvision import models
    from aimet_torch import utils

    model = models.resnet18(pretrained=True).eval()

    # Replace any ReLU6 layers with ReLU , model is updated in-place
    utils.replace_modules_of_type1_with_type2(model, 
                                              torch.nn.ReLU6,
                                              torch.nn.ReLU)
    \end{lstlisting}    
\end{itemize}

\section{Quantization range setting}
\label{sec:range_setting}
A fundamental step in the PTQ process is finding suitable quantization ranges for each quantizer. Quantization range setting refers to the method of determining clipping thresholds of the quantization grid, $q_{\text{min}}$ and $q_{\text{max}}$ (see equation~\ref{eq:quant_function}). The range setting method influences the trade-off between between clipping and rounding error described in section \ref{sec:quant_scheme} and consequently  the impact on the final task loss. Each of the methods described here provides a different trade-off between the two quantities. These methods typically optimize local cost functions instead of the task loss. This is because in PTQ we aim for computationally fast methods without the need for end-to-end training.

Weights can usually be quantized without any need for calibration data. However, determining parameters for activation quantization often requires a few batches of calibration data. Typically, $500$ to $1{\small,}000$ data samples have shown good results for vision tasks. Note that quantization range setting is required regardless of the PTQ optimization technique employed, e.g CLE/BC or AdaRound.

\paragraph{\textbf{Min-max}} To cover the whole dynamic range of the tensor, we can define the quantization parameters as follows 
\begin{align}
    q_{\text{min}} &= \min \mat{V}, \\
    q_{\text{max}} &= \max \mat{V},
\end{align}
where $\mat{V}$ denotes the tensor to be quantized. This leads to no clipping error. 
However, this approach is sensitive to outliers, as strong outliers may cause excessive rounding errors.

\paragraph{\textbf{Signal-to-quantization-noise ratio (SQNR)}}
One way to alleviate the issue of large outliers is to use Signal-to-Quantization-Noise (SQNR) based range setting. SQNR approach is similar
to the Mean Square Error (MSE) minimization approach known in the literature. In SQNR range setting method we find $q_\text{min}$ and $q_\text{max}$ that minimize the total MSE between the original and the quantized tensor,
where the error coming from the quantization noise and saturation is differently weighted.


\paragraph{\textbf{Code example}}
\begin{itemize}
 \item [\colorbox{gray!40}{API}]
    Quantization simulation on AIMET allows user to choose the quantization range setting scheme. It supports two options - \texttt{QuantScheme.post\_training\_tf} that uses standard min-max and \texttt{QuantScheme.post\_training\_tf\_enhanced} uses SQNR technique described above.
    \begin{lstlisting}[caption={Quantization range setting API example.},language=Python]

    import torch
    from aimet_torch.examples import mnist_torch_model
    # Quantization related import
    from aimet_torch.quantsim import QuantizationSimModel

    model = mnist_torch_model.Net().to(torch.device('cuda'))

    # create Quantization Simulation model with appropriate quant_scheme, supported schemes :
    # QuantScheme.post_training_tf_enhanced
    # QuantScheme.post_training_tf
    
    quant_scheme = QuantScheme.post_training_tf_enhanced
    sim = QuantizationSimModel(model, 
              dummy_input=torch.rand(1, 1, 28, 28),
              quant_scheme=quant_scheme,
              default_output_bw=8,
              default_param_bw=8)
    \end{lstlisting}
\end{itemize}

\section{Bias correction}
\label{sec:bias_correction}
Another common issue is that quantization error is often biased. This means that the expected output of the original and quantized layer or network is shifted ($\eop{\mat{W}\vec{x}} \neq \mathbb{E}[\matq{W}\vec{x}]$). This kind of  error is more pronounced in depth-wise separable layers with only a few weights per output channel (usually 9 for a $3 \small{\times} 3$ kernel). The main contributor to this error is often the clipping error, as a few strongly clipped outliers will likely lead to a shift in the expected distribution.

By adapting the layer's bias parameter, we can correct for the bias in the noise. This correction recovers part of the accuracy of the original FP32 model. This correction term is a vector with the same shape as the layer bias and can thus be absorbed into the bias without any additional overhead at inference. AIMET supports two methods for bias correction: \textit{empirical bias correction} and \textit{analytic bias correction}.

\paragraph{\textbf{Empirical bias correction}}
If we have a calibration dataset, then the bias correction term can simply be calculated by comparing the activations of the quantized and full precision model.
\paragraph{\textbf{Analytic bias correction}}
\citet{dfq} introduced a method to analytically correct the biased error without the need for data. For common networks with batch normalization and ReLU activations, they use the batch normalization statistics of the preceding layer to approximate the expected input distribution $\eop{\vec{x}}$ and this correct for the bias error analytically.

\paragraph{\textbf{Code example}}
\begin{itemize}
    \item [\colorbox{gray!40}{API}]
    AIMET performs empirical bias correction by default. 

    \begin{lstlisting}[caption={Bias correction API example.},language=Python]

    from torchvision import models
    from aimet_torch.bias_correction import correct_bias
    from aimet_torch.quantsim import QuantParams

    model = models.resnet18(pretrained=True).eval()
    params = QuantParams(weight_bw=4, act_bw=4,
                         round_mode="nearest", 
                         quant_scheme='tf_enhanced')

    # Perform Bias Correction
    # perform_only_empirical_bias_corr to True to 
    # enforce only empirical bias correction
    correct_bias(model.to(device="cuda"),
                 params=params,
                 num_quant_samples=1000,
                 data_loader,
                 num_bias_correct_samples=512,
                 perform_only_empirical_bias_corr)
    \end{lstlisting}

    \item [\colorbox{gray!40}{Usage Note}]   
    \item User can choose the type of bias correction to be applied using the \texttt{perform\_only\_empirical\_bias\_corr} flag in the API below. If set to \texttt{True}, only empirical bias correction is applied. If time is not a bottleneck, then it is suggested to use only empirical bias correction.
    \item In case of analytic bias correction (\texttt{perform\_only\_empirical\_bias\_corr = False}), AIMET automatically detects candidate convolution layers with associated batch normalization statistics in a given model to perform bias correction. For layers without associated batch normalization statistics, empirical bias correction is applied if a calibration dataset is provided.
    \item Bias correction requires a set of training or calibration data. We have observed that providing $500-1,000$ samples has worked well for vision tasks.
\end{itemize}

\section{AdaRound}
\label{sec:adaround}
\begin{figure}[ht]
    \centering
    \includegraphics[width=.60\textwidth]{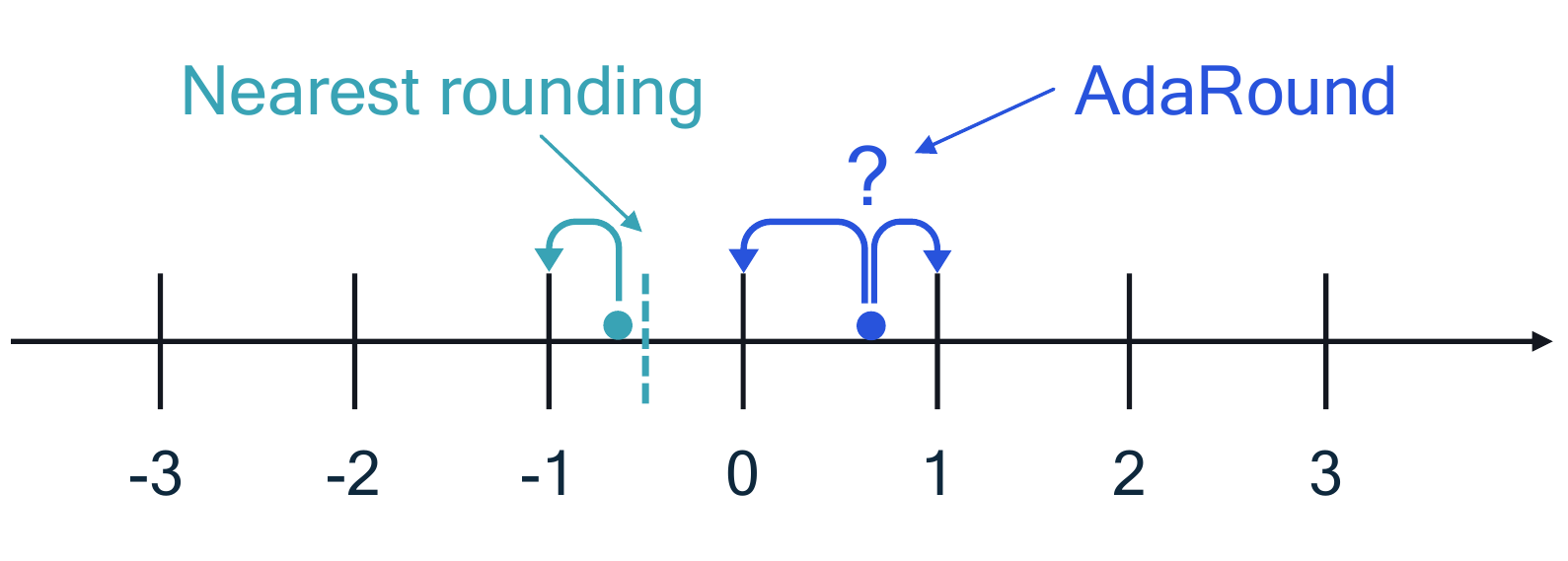}
    \caption{AdaRound versus round-to-nearest approach.}
    \label{fig:adaround_basic}
\end{figure}

Neural network weights are usually quantized by projecting each FP32 value to the \textit{nearest} quantization grid point, as indicated by $\round*{\cdot}$ in equation \eqref{eq:quant_operation} for a uniform quantization grid. We refer to this quantization strategy as \textit{rounding-to-nearest}. 
The way we round weights during the quantization operation has a significant impact on the performance of the network.  
AdaRound (adaptive rounding) provides a theoretically sound, computationally fast method for rounding weights. It requires only a small number of unlabeled data samples (example : $500$ - $2{\small,}000$ samples were used on vision tasks), no hyperparameter tuning or end-to-end finetuning, and can be applied to fully connected and convolutional layers of any neural network. AdaRound optimizes a local loss function using the unlabeled training or calibration data to adaptively decide whether to quantize a specific weight to the integer value below or above. Figure \ref{fig:adaround_basic} illustrates schematically the difference between round-to-nearest and AdaRound. During quantization simulation (cf. section \ref{sec:quantsim_intro}), AIMET by default rounds the weights of a layer to the nearest value. However, for performance enhancement especially for low bit-width integer quantization, AdaRound should be invoked.

\newpage
\paragraph{\textbf{Code example}}
\begin{itemize}
    \item [\colorbox{gray!40}{API}]
    AIMET supports one generic API to perform AdaRound on a given model as shown below. It returns an updated model with optimized weights and saves the corresponding quantization encodings in a JSON output file, at the user specified path defined by the \texttt{encoding\_path} argument.
    \begin{lstlisting}[caption={AdaRound API example.},language=Python]

    from torchvision import models
    from aimet_torch.quantsim import QuantizationSimModel
    from aimet_torch.adaround.adaround_weight import Adaround, AdaroundParameters

    torch.cuda.empty_cache()
    
    model = models.resnet18(pretrained=True).eval()
    model = model.to(torch.device('cuda'))
    input_shape = (1, 3, 224, 224)
    dummy_input = torch.randn(input_shape).to(torch.device('cuda'))

    # configuration parameters
    params = AdaroundParameters(data_loader, num_batches=4,
                                default_num_iterations=50,
                                default_reg_param=0.01,
                                default_beta_range=(20, 2))

    # Returns a model with adarounded weights and corresponding encodings
    quant_scheme = QuantScheme.post_training_tf_enhanced
    adarounded_model = Adaround.apply_adaround(model, 
                           dummy_input, 
                           params, 
                           path='./',
                           filename_prefix='resnet18', 
                           default_param_bw=8,
                           default_quant_scheme=quant_scheme,
                           default_config_file=None)

    # Create quant-sim using adarounded_model    
    sim = QuantizationSimModel(adarounded_model, dummy_input
              quant_scheme=quant_scheme,
              default_param_bw=param_bw,
              default_output_bw=output_bw)

    # Set and freeze encodings to use same quantization grid and then invoke compute encodings
    sim.set_and_freeze_param_encodings(
              encoding_path='./resnet18.encodings')

    sim.compute_encodings(forward_pass_callback=send_samples, 
              forward_pass_callback_args=None)
    \end{lstlisting}

    \item [\colorbox{gray!40}{Usage Note}]
        \item The AdaRound optimization has several hypermarameters that are exposed to the user. The default values in AIMET provide good and stable results based on our experimentation. If the default hyperparameters lead to unsatisfactory accuracy or long run-times, consider the following guidelines on how to tune them:
        \begin{itemize}            
            \item The most influential hyperparameters are the \textit{number of batches} and the \textit{number of iterations} (default $10{\small,}000$). These influence the amount of data used and the duration of the optimization, respectively. We recommend using at least $500$ to $2{\small,}000$ samples, that has worked well for vision tasks.(But, would need to be determined on a case-by-case basis) (batch size of $64$ and number of batches $16$ correspond to $1{\small,}024$ samples).
            \item The rest of the hyperparameters (\texttt{reg\_param}, \texttt{beta\_range}, \texttt{warm\_start}) are related to the AdaRound formulation and do not need to be changed for most cases. If tuning is required, please consult the AdaRound paper \citep{adaround} or an expert. 
        \end{itemize}                
            \item When creating a QuantizationSimModel using the post-AdaRound model, use the QuantizationSimModel provided API for setting and freezing parameter encodings before computing the encodings. Note, that freezing of encodings is essential to realize AdaRound performance benefits, because the weight adaptation assumes an underlying quantization grid and this must be used during inference as well.
           
\end{itemize}

\section{Results}
\label{sec:aimet_ptq_results}
In this section, we demonstrate the effectiveness of the post-training techniques discussed in this chapter. Table \ref{tbl:ptq_results_aimet} shows the INT8 quantized accuracy for popular use cases, such as object classification and semantic segmentation using CLE/BC techniques (detailed in section \ref{sec:CLE}). We can see that with CLE/BC 8-bit quantization, the accuracy is within 1\% of original FP32 model, thus, demonstrating CLE/BC's effectiveness without any model fine-tuning. 
\\
\input{tables/aimet_ptq_results}

Table \ref{tbl:adaround_adas_aimet} demonstrates the quantized accuracy on an object detection model for Advanced Driver-Assistance System (ADAS) application. It is challenging to quantize this model despite the use of CLE/BC. Quantized accuracy with traditional nearest rounding approach is significantly inferior compared to FP32 performance. However, with AdaRound, we can achieve INT8 accuracy within 1\% of FP32.  
\input{tables/03_adaround_adas_aimet}

\input{sections/046_ptq_diagnostics}

%% file: tables/aimet_ptq_results.tex
\begin{table}[h]
    \centering
  
    \begin{tabular}{ l | c| c | c }
    \toprule
         Model &  Baseline (FP32) &  W8/A8 without CLE/BC &  AIMET W8/A8 with CLE/BC\\
    \midrule
     MobileNetV2  &  71.72\% &  0.09\% &  71.08\% \\
     ResNet-50  & 76.05\% & 75.42\%  & 75.45\% \\
     DeepLabV3 & 72.65\% & 47.02\% & 71.91\% \\
    \bottomrule
    \end{tabular}
    
    \caption{Results on ImageNet (top-1 accuracy) with AIMET PTQ methods: CLE and bias correction.}
     \label{tbl:ptq_results_aimet}
\end{table}

%% file: tables/03_adaround_adas_aimet.tex
\begin{table}[h]
    \centering
  
    \begin{tabular}{ l | c| c | c }
    \toprule
         Model &  Baseline (FP32) & Round-to-nearest &  AIMET AdaRound\\
    \midrule
     ADAS object detect (mAP)  & 82.20\% &  49.85\% &  81.21\% \\    
    \bottomrule
    \end{tabular}
    
    \caption{AdaRound on ADAS object detection model. mAP = mean average precision. \\
    Round-to-nearest and AdaRound with W8/A8 quantization configuration.}
     \label{tbl:adaround_adas_aimet}
\end{table}


%% file: sections/046_ptq_diagnostics.tex
 \section{Debugging}
\label{sec:debugging}

\begin{figure}[p]
    \centering
    \includegraphics[width=1\textwidth]{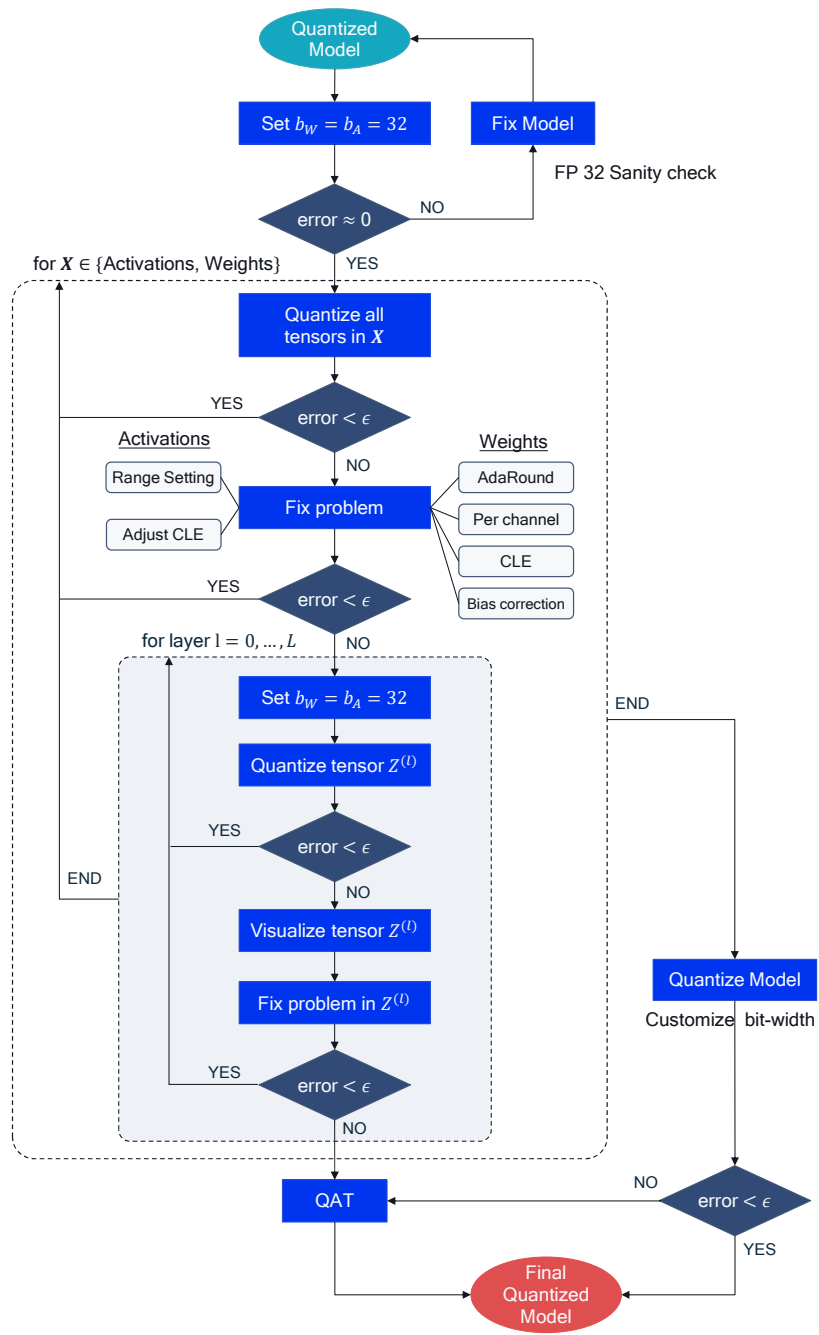}
    \caption{PTQ debugging flow chart. Error is the difference between floating-point and quantized model accuracy. }
    \label{fig:ptw_debugging_flowchart}
\end{figure}

We showed that the standard PTQ pipeline can achieve competitive results for a wide range of models and networks. However, if after following the steps of our pipeline, the model's performance is still not satisfactory, we recommend a set of diagnostics steps to identify the bottlenecks and improve the performance. While this is not strictly an algorithm, these debugging steps can provide insights on why a quantized model underperforms and help to tackle the underlying issues. These steps are shown as a flow chart in figure \ref{fig:ptw_debugging_flowchart} and are described in more detail below:

\begin{description}
    \item[FP32 sanity check] An important initial debugging step is to ensure that the floating-point and quantized model behave similarly in the forward pass, especially when using custom quantization pipelines. Set the quantized model bit-width to 32 bits for both weights and activation, or by-pass the quantization operation, if possible, and check that the accuracy matches that of the FP32 model.
    
    \item[Weights or activations quantization] The next debugging step is to identify how activation or  weight quantization impact the performance independently. Does performance recover if all weights are quantized to a higher bit-width while activations are kept in a lower bit-width, or conversely if all activations use a high bit-width and activations a low bit-width? This step can show the relative contribution of activations and weight quantization to the overall performance drop and point us towards the appropriate solution.

    \item[Fixing weight quantization] If the previous step shows that weight quantization does cause significant accuracy drop, then there are a few solutions to try:
    \begin{itemize}
        \item Apply CLE described in section \ref{sec:CLE}, especially for models with depth-wise separable convolutions.
        \item Apply bias correction (cf. section \ref{sec:bias_correction}) or AdaRound (cf. section \ref{sec:adaround}), if calibration data is available.
    \end{itemize}
     \item[Fixing activation quantization] To reduce the quantization error from activation quantization, we can also try using different range setting methods or adjust CLE (cf. section \ref{sec:CLE}) to take activation quantization ranges into account, as vanilla CLE can lead to uneven activation distribution.
    \item[Per-layer analysis] If the global solutions have not restored accuracy to acceptable levels, we consider each quantizer individually. We set each quantizer sequentially, to the target bit-width while keeping the rest of the network to 32 bits (see inner for loop in figure \ref{fig:ptw_debugging_flowchart}). 
    
    \item[Visualizing layers] If the quantization of a individual tensor leads to significant accuracy drop, we recommended visualizing the tensor distribution at different granularities, and dimensions, e.g., per-token or per-embedding for activations in BERT. Please refer to \href{https://quic.github.io/aimet-pages/index.html}{AIMET documentation} for details on visualization APIs supported by AIMET.
 
    \item[Fixing individual quantizers] The visualization step can reveal the source of the tensor's sensitivity to quantization. Some common solutions involve custom range setting for this quantizer or allowing a higher bit-width for problematic quantizer. If the problem is fixed and the accuracy recovers, we continue to the next quantizer. If not, we may have to resort to other methods, such as quantization-aware training (QAT), covered in chapter \ref{ch:QAT}.

\end{description}

After completing the above steps, the last step is to quantize the complete model to the desired bit-width. If the accuracy is acceptable, we have our final quantized model ready to use. Otherwise, we can consider higher bit-widths and smaller granularities or revert to more powerful quantization methods, such as quantization-aware training.

 \paragraph{\textbf{Code example}}
    \begin{itemize}
    \item [\colorbox{gray!40}{API}]
    AIMET quantization simulation, covered in section \ref{sec:quantsim_intro}, allows the user to specify bit-widths for activations and parameters of a given model. The code examples below demonstrate a way of specifying model-level configuration of bit-width for activation and parameters. In addition, the user can also specify custom rules to be used for quantization of the model using a runtime configuration file (specifically, the params and op\_type knobs in the configuration file detailed in section \ref{sec:quantsim_config} can be used for this purpose).
    \begin{lstlisting}[caption={User-configurable activation and parameter bit-width quantization simulation API example.},language=Python]
    
    import torch
    from aimet_torch.examples import mnist_torch_model
    # Quantization related import
    from aimet_torch.quantsim import QuantizationSimModel

    model = mnist_torch_model.Net().to(torch.device('cuda'))

    # Create a quantization simulation model
    # Adds simulation ops, configures these simulation ops

    # Customize bit-width using:
    # default_output_bw set to desired bit-width to be used for activation quantization
    # and default_param_bw set to desired bit-width to be used for parameter quantization.
    # Additionally, config_file option could be used to specify custom runtime bit-width configurations.

    sim = QuantizationSimModel(model, 
              dummy_input=torch.rand(1, 1, 28, 28),
              default_output_bw=8, 
              default_param_bw=8,
              config_file='custom_runtime_config.json')

    # Find optimal quantization parameters (like scale/offset)
    # The callback method should send representative data 
    # samples through the model. The forward_pass_callback_args 
    # argument is passed as-is to the callback method
    
    sim.compute_encodings(forward_pass_callback=send_samples, 
              forward_pass_callback_args=None)

    # Optionally, the user can export out the quantization 
    # parameters (like per-layer scale/offset)
    sim.export(path='./',
               filename_prefix='quantized_mnist', 
               dummy_input=torch.rand(1, 1, 28, 28))
    \end{lstlisting}
\end{itemize}

%% file: sections/05_qat.tex
\chapter{Quantization-Aware Training}
\label{ch:QAT}
\section{Introduction}
The post-training quantization techniques covered in the previous chapters are the first go-to tool in our quantization toolkit. They are very effective and fast to implement because they do not require retraining the network with labeled data. However, for some models or use cases, they may be insufficient. Post-training techniques may not be enough to mitigate the significant quantization error incurred by low-bit quantization. In these cases, we resort to \textit{quantization-aware training} (QAT). QAT models the quantization noise during training and allows the model to find better solutions than post-training quantization. However, the higher accuracy comes with the usual costs of neural network training, i.e. longer training times, need for labeled data and hyperparameter search. In this chapter, we briefly explore how back-propagation works in networks with simulated quantization and provide a standard pipeline for training models with QAT effectively.

\begin{figure}[ht]
    \centering
    \includegraphics[width=0.5\textwidth]{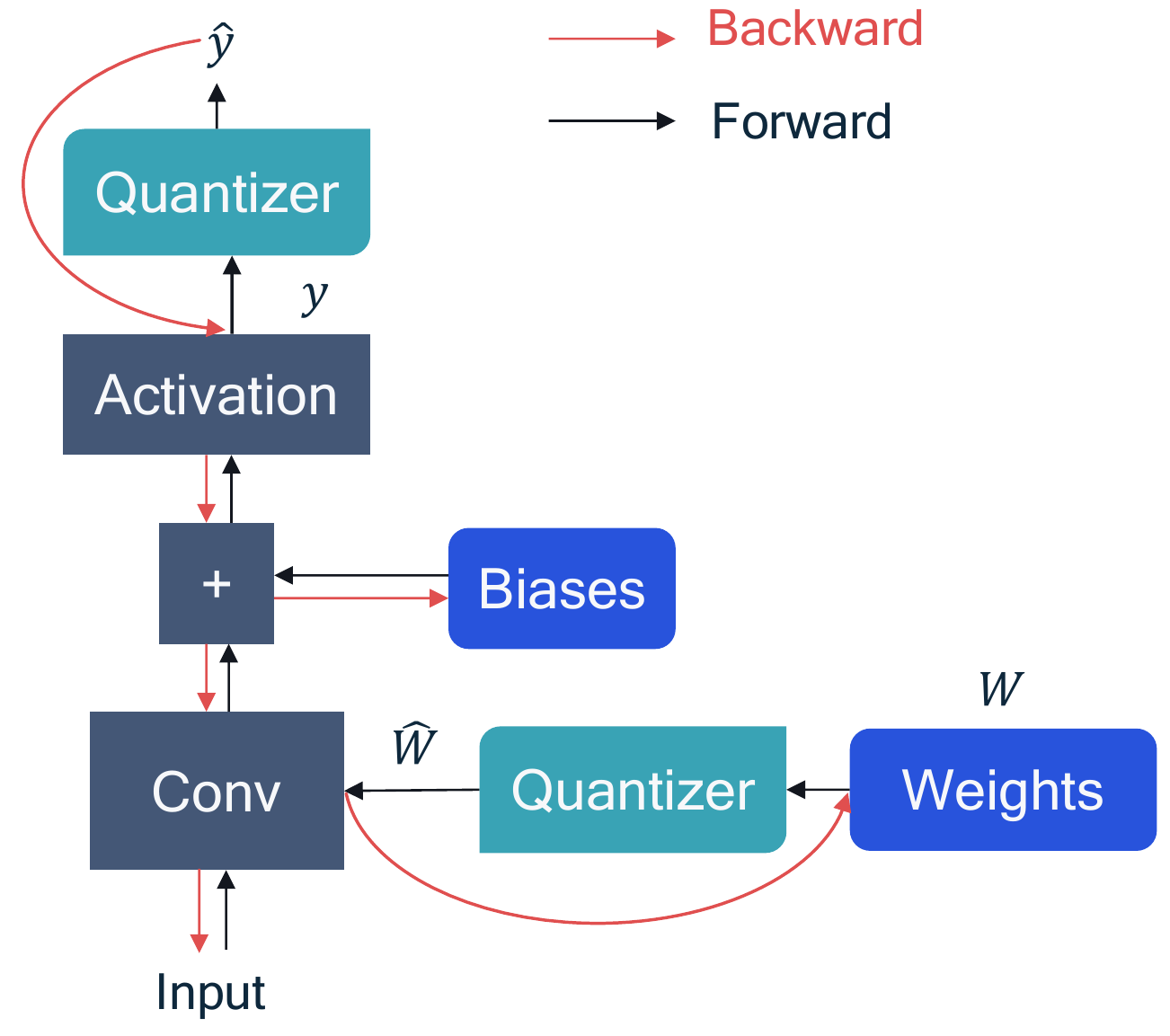}
    \caption{Forward and backward computation graph for quantization aware training with STE assumption.}
    \label{fig:sim_quant_backward}
\end{figure}

In section \ref{sec:quantsim_intro}, we saw how quantization can be simulated using floating-point in deep learning frameworks. However, if we look at the computational graph of figure \ref{fig:simulation}, to train such a network we need to back-propagate through the simulated quantizer block. This poses an issue because the gradient of the round-to-nearest operation in equation \eqref{eq:quant_operation} is either zero or undefined everywhere, which makes gradient-based training impossible. A way around this would be to approximate the gradient using the \textit{straight-through estimator} (STE,~\citealt{bengio2013estimating}), which approximates the gradient of the rounding operator as straight-through estimator.
Using this gradient definition we can now back-propagate through the quantization blocks. Figure \ref{fig:sim_quant_backward} shows a simple computational graph for the forward and backward pass used in quantization-aware training. The forward pass is identical to that of figure~\ref{fig:simulation}, but in the backward pass we effectively skip the quantizer block due to the STE assumption.The quantization ranges for weights and activations can be updated at each iteration most commonly using the min-max range \citep{krishnamoorthi} or ranges can also be learned like in LSQ \citep{lsq}.


\section{QAT pipeline}
\begin{figure}[h]
    \centering
    \includegraphics[width=1.0\textwidth]{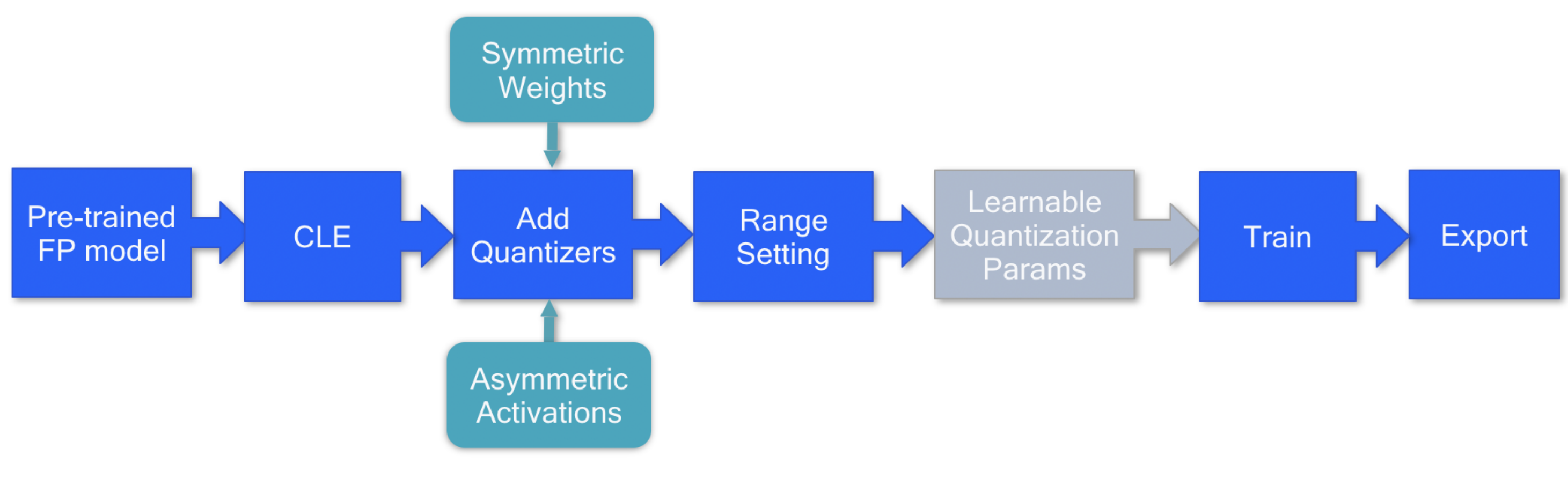}
    \caption{Quantization-aware training pipeline. The blue boxes represent the steps and the turquoise boxes recommended choices and grey box is an optional step.}
    \label{fig:qat_pipeline}
\end{figure}

In this section, we present a best-practice pipeline for QAT. We illustrate the recommended pipeline in figure \ref{fig:qat_pipeline}. This pipeline yields good QAT results over a variety of computer vision and natural language processing models and tasks, and can be seen as the go-to tool for achieving low-bit quantization performance. As discussed in the previous sections, we always start from a pre-trained FP32 model and follow some PTQ steps in order to have faster convergence and higher accuracy.

\begin{description}
\item[Cross-layer equalization] Similar to PTQ, first we apply CLE to the full precision model. As covered in section \ref{ch:PTQ}, this step is necessary for models that suffer from imbalanced weight distributions, such as MobileNet architectures. For other networks or in the case of per-channel quantization this step is optional.

\item[Add quantizers] Next, we choose the quantizers and add quantization operations to the network as described in section \ref{sec:quantsim_intro}. The choice for quantizer might depend on the specific target hardware, for common AI accelerators we recommend using symmetric quantizers for the weights and asymmetric quantizers for the activations (refer, section \ref{sec:practical considerations}). If supported by the HW/SW stack, then it is favorable to use per-channel quantization for weights. At this stage we will also take care that our simulation of batch normalization is correct, as discussed in section \ref{sec:quantsim_bn_fold}. AIMET finds quantization parameters, such as scale/offsets, for these inserted quantization simulations operations. To this end, AIMET requires the user to provide a callback method that feeds a few representative data samples through the model. These samples can either be from the training or calibration datasets. Generally, samples in the order of $1,000$ have been sufficient for AIMET to find optimal quantization parameters (vision tasks).

\item[Range setting] Before training we have to initialize all the quantization parameters. A better initialization will help faster training and might improve the final accuracy, though often the improvement is small. In general, we recommend to set all quantization parameters using the layer-wise SQNR based criteria. In the specific case of per-channel quantization, using the min-max setting can sometimes be favorable.

\item[Learnable Quantization Parameters]
Optionally, the quantizer parameters could be made learnable. Learning the quantization parameters directly, rather than
updating them at every epoch, leads to higher performance especially when dealing with low-bit quantization (for more details on learnable quantizer, please refer to our white paper on neural network quantization \citep{whitepaper}).

\item[Train] Quantization simulation model is returned from AIMET (refer section \ref{sec:quantsim_intro}, that the we can use as a drop-in replacement for the original model in the training pipeline to perform quantization-aware training.

\item[Export] After fine-tuning, the user can export the optimized quantized model using the API detailed in section \ref{sec:quantsim_export}.

\end{description}


\paragraph{\textbf{Code example}}
\begin{itemize}
\item [\colorbox{gray!40}{API}]
    Shown below is an example for using AIMET Quantization simulation for QAT.
    \begin{lstlisting}[caption={Quantization-aware training API example.},language=Python]   
    
    import torch
    from aimet_torch.examples import mnist_torch_model
    # Quantization related import
    from aimet_torch.quantsim import QuantizationSimModel

    model = mnist_torch_model.Net().to(torch.device('cuda'))

    # create Quantization Simulation model
    sim = QuantizationSimModel(model, 
              dummy_input=torch.rand(1, 1, 28, 28),                               
              default_output_bw=8, 
              default_param_bw=8)

    # Quantize the untrained MNIST model
    sim.compute_encodings(
              forward_pass_callback=send_samples, 
              forward_pass_callback_args=5)

    # Fine-tune the model's parameter using training
    trainer_function(model=sim.model, epochs=1, 
                     num_batches=100, use_cuda=True)

    # Export the model and corresponding quantization encodings
    sim.export(path='./', filename_prefix='quantized_mnist', 
               dummy_input=torch.rand(1, 1, 28, 28))               
    \end{lstlisting}
\end{itemize}

\begin{itemize}
\item [\colorbox{gray!40}{Usage Note}]
Here, we present general guidelines for improving the quantized model's accuracy and convergence time for quantization-aware training:

\begin{description}
\item[Initialization] Start from a well initialized quantized model. Apply post-training quantization techniques, such as CLE before training.
\item[Hyperparameters] \hspace{1pt}
\begin{itemize}
    \item \textit{Number of epochs}: $10$-$20$\% of the original epochs are generally sufficient for convergence. For example: With typical vision models, we have observed $15$-$20$ epochs yield good results.
    \item\textit{Learning rate}: Comparable (or one order higher) to FP32 model’s final learning rate at convergence. Results in AIMET are with learning of the order $1\small{e}$-$6$.
    \item \textit{Learning rate schedule}: Divide learning rate by $10$ every $5$-$10$ epochs.
    \item \textit{Optimizer config}: When using learnable quantizer, we recommend using Adam/RMS Prop optimizer for quantization parameters, while following other recommendations for the weight fine-tuning.
\end{itemize}
\end{description}

%
\end{itemize}

\subsection{Batch normalization folding and QAT} \label{sec:qat_bn_folding}
In section \ref{sec:batch_norm_folding}, we introduced batch normalization folding that absorbs the scaling and addition into a linear layer to allow for more efficient inference. During quantization-aware training, we want to simulate inference behavior closely, which is why we have to account for batch normalization folding during training. Note that in some QAT literature, the batch normalization folding effect is ignored, and batch normalization is left out of the quantization operations. This does not matter for per-channel quantization because, during inference, the batch normalization layers apply a fixed per-channel rescaling that can be folded into the per-channel quantization scale factors. However, for per-tensor quantization, if we fold batch normalization into the weight tensor only during deployment we may incur significant accuracy drop. This is because the network has been train to adapt to different quantization noise.

A simple but effective approach to modeling batch normalization folding in QAT is to \textit{statically fold} the batch normalization scale and offset into the linear layer's weights and bias before training. This corresponds to a re-parametrization of the weights and effectively removes the batch normalization operation from the network entirely. AIMET uses static batch normalization folding approach.


\section{Results}
\label{sec:QAT_Results}

\input{tables/aimet_qat}
\input{tables/aimet_qat_rnn}

Using our QAT pipeline, we quantize and evaluate the models we used for PTQ (cf. section \ref{sec:aimet_ptq_results}). Our results are presented in table \ref{tbl:aimet_qat} for two different model architectures - MobileNetV2 and ResNet50. AIMET QAT narrows the gap relative to FP32 accuracy (<0.5\% delta for MobileNetV2).

In addition, AIMET also supports quantization simulation and quantization-aware training for recurrent neural networks, such as RNN, LSTM and GRU. Using QAT with AIMET, a DeepSpeech2 model with bi-directional LSTMs can be quantized to 8-bit precision with  minimal drop in accuracy, as shown in table \ref{tbl:aimet_qat_rnn_t}.

%% file: tables/aimet_qat.tex
\begin{table}[!h]
    \centering

    \begin{tabular}{ l| c | c | c}
    \toprule
         Model & Baseline (FP32) & AIMET PTQ & AIMET QAT\\
    \midrule
     MobileNetV2 & 71.72\% & 71.08\% & 71.23\% \\
     ResNet50 &76.05\% & 75.45\% & 76.44\% \\
    \bottomrule
    \end{tabular}    

    \caption{AIMET quantization-aware training results for ImageNet (top-1 accuracy).
     Both PTQ and QAT use W8/A8 quantization config. AIMET QAT is performed with PTQ initialization.}     
     \label{tbl:aimet_qat}
\end{table}

%% file: tables/aimet_qat_rnn.tex
\begin{table}[!h]
    \centering

    \begin{tabular}{ l | c | c }
    \toprule
         Model &  Baseline (FP32) & AIMET QAT\\
    \midrule
     DeepSpeech2 (WER) &  9.92\% & 10.22\%  \\

    \bottomrule
    \end{tabular}    

    \caption{Word Error Rate (WER) result for DeepSpeech2 with bi-directional LSTMs using AIMET quantization-aware training (the lower the better).
    AIMET QAT uses W8/A8 quantization config and PTQ initialization.}     
     \label{tbl:aimet_qat_rnn_t}
\end{table}

%
%



%% file: sections/06_conclusions.tex
\chapter{Summary and Conclusions}
Deep learning has become an integral part of many machine learning applications and can now be found in countless electronic devices and services, from smartphones and home appliances to drones, robots and self-driving cars. As the popularity and reach of deep learning in our everyday life increases, so does the need for fast and power-efficient neural network inference. Neural network quantization is one of the most effective ways of reducing the energy and latency requirements of neural networks during inference.

Quantization allows us to move from floating-point representations to a fixed-point format and, in combination with dedicated hardware utilizing efficient fixed-point operations, has the potential to achieve significant power gains and accelerate inference. However, to exploit these savings, we require robust quantization methods that can maintain high accuracy, while reducing the bit-width of weights and activations. 
To this end, we consider two main classes of quantization algorithms: Post-Training Quantization (PTQ) and Quantization-Aware Training (QAT). 

Post-training quantization techniques take a pre-trained FP32 networks and convert it into a fixed-point network without the need for the original training pipeline. This makes them a lightweight, push-button approach to quantization with low engineering effort and computational cost. We describe a series of recent advances in PTQ and introduce a PTQ pipeline that leads to near floating-point accuracy results for a wide range of models and machine learning tasks. In particular, using the proposed pipeline we can achieve 8-bit quantization of weights and activations within only 1\% of the floating-point accuracy for various networks (vision tasks). In addition, we introduce a debugging workflow to effectively identify and fix problems that might occur when quantizing new networks.

Quantization-aware training models the quantization noise during training through simulated quantization operations. This training procedure allows for better solutions to be found compared to PTQ. Similar to PTQ, we introduce a standard QAT pipeline. 

Both PTQ and QAT appproaches are part of Qualcomm's AI Model Efficiency toolkit (AIMET). As demonstrated through code examples in this paper, AIMET offers simple, easy-to-use APIs to apply these techniques on PyTorch models. Additionally, AIMET also supports TensorFlow models and detailed code examples for the same are available on the open source GitHub page - \href {https://quic.github.io/aimet-pages/index.html}{https://quic.github.io/aimet-pages/index.html}.  The performance of AIMET's techniques has been validated on a wide range of models and tasks. These optimized models are also made available on GitHub as AIMET Model Zoo [\href{https://github.com/quic/aimet-model-zoo}{https://github.com/quic/aimet-model-zoo}]. We hope that our proposed pipelines and corresponding AIMET code examples shared through this white paper will help engineers deploy high-performing quantized models with less time and effort.

%% file: sections/07_aimet_spec.tex
\chapter{Appendix}
\section{Glossary}
\input{tables/glossary}

\section{Supported layer types}
\label{sec:supported_types}
\subsection{PyTorch}
\begin{itemize}    
    \item [{\colorbox{gray!40}{Convolution Layers}}]
    torch.nn.Conv1d, torch.nn.Conv2d, torch.nn.ConvTranspose1d, torch.nn.ConvTranspose2d
    \item [{\colorbox{gray!40}{BatchNorm Layers}}]  
    torch.nn.BatchNorm1d, torch.nn.BatchNorm2d
\end{itemize}

\section{PyTorch model guidelines}
\label{sec:pt_model_guidelines}
\begin{itemize}
\item In order to make full use of AIMET features, there are several guidelines users are encouraged to follow when defining PyTorch models.
\item Model should support conversion to ONNX. The user could check compatibility of model for ONNX conversion as shown below:
    \begin{lstlisting}[caption={ONNX check},language=Python]   
    
    model = Model()
    torch.onnx.export(model, <dummy_input>, <onnx_file_name>):
    \end{lstlisting}
\item Define layers as modules instead of using \texttt{torch.nn.functional} equivalents, when using activation functions and other stateless layers. PyTorch allows the user to either define the layers as modules (instantiated in the constructor and used in the forward pass), or use a \texttt{torch.nn.functional} equivalent purely in the forward pass. For AIMET requires that all layers are defined as modules to successfully add quantization simulation operations. Changing the model definition to use modules instead of functionals, is mathematically equivalent and does not require the model to be retrained. This many not be possible in certain cases, where operations can only be represented as functionals and not as class definitions, but should be followed whenever possible.

\item Avoid reuse of class defined modules. Modules defined in the class definition should only be used once. If any modules are being reused, instead define a new identical module in the class definition.
\end{itemize}

\subsection{Other layers and quantization}
\label{sec:other_layers_and_quantization}
There are many other types of layers being used in neural networks. How these are modeled depends greatly on the specific hardware implementation. Sometimes the mismatch between simulated quantization and on-target performance is down to layers not being properly quantized. Here, we provide some guidance on how to simulate quantization for a few commonly used layers:
\begin{description}
    \item[Max pooling] Activation quantization is not required because the input and output values are on the same quantization grid.
    \item[Average pooling] The average of integers is not  necessarily an integer. For this reason, a quantization step is required after average-pooling. However, we use the same quantizer for the inputs and outputs as the quantization range does not significantly change.
    \item[Element-wise addition] Despite its simple nature, this operation is difficult to simulate accurately. During addition, the quantization ranges of both inputs have to match exactly. If these ranges do not match, extra care is needed to make addition work as intended. There is no single accepted solution for this but adding a requantization step can simulate the added noise coarsely. Another approach is to optimize the network by using the same quantization parameters for all inputs to the addition. This would prevent the requantization step but may require fine-tuning. 
    \item[Concatenation] The two branches that are being concatenated generally do not share the same quantization parameters. This means that their quantization grids may not overlap making a requantization step necessary. As with element-wise addition, it is possible to optimize your network to have shared quantization parameters for the branches being concatenated. 
\end{description}

%% file: tables/glossary.tex
\begin{table}[h]
  \centering
  \begin{tabular}{@{}cc@{}}
    \toprule
    Abbrevation  &  Definition  \\
    \midrule
    AdaRound & Adaptive Rounding \\
    AIMET    & AI Model Efficiency Toolkit \\
    API      & Application Programming Interface \\
    BNF      & Batch Normalization Folding \\
    CLE      & Cross-layer Equalization\\
    DFQ      & Data-free Quantization\\
    DLC      & Deep Learning Container\\
    PTQ      & Post-training Quantization \\
    QAT      & Quantization-aware Training\\
    QuantSim & Quantization Simulation \\
    SDK      & Software Development Kit\\
    \bottomrule
  \end{tabular}
  \caption{Abbreviations}
\end{table}